\documentclass{article}

\usepackage{times}
\usepackage{wrapfig}
\usepackage{amssymb}
\usepackage{amsmath}
\usepackage{amsthm}
\usepackage[hyphens]{url}
\usepackage{hyperref}
\usepackage[capitalize,noabbrev]{cleveref}
\usepackage{xcolor}
\usepackage{subcaption}
\usepackage[title]{appendix}
\usepackage{multirow}
\usepackage{graphicx}
\usepackage{xspace}
\usepackage{booktabs} 
\usepackage[subtle]{savetrees}
\usepackage{algorithm}
\usepackage{algorithmicx}
\usepackage{algpseudocodex}
\usepackage{longtable}
\usepackage{microtype}

\newcommand{\sys}{LCPO\xspace}
\newcommand{\bigsys}{Locally Constrained Policy Optimization\xspace}

\renewcommand{\cite}[1]{\citep{#1}}

\usepackage{tikz}
\usetikzlibrary{calc}

\newcommand{\shadeline}[1]{%
    \begin{tikzpicture}[remember picture, overlay]
     \tikz[overlay,remember picture,baseline] \node [anchor=base] (begin) {};
        \fill[gray, opacity=0.10] ($(begin|-begin)+(-#1,+7.5pt)$) rectangle ($(begin|-begin)+(\dimexpr\labelwidth+\labelsep+\linewidth-#1\relax,-3.5pt)$);
     \end{tikzpicture}%
}
\algrenewcommand\alglinenumber[1]%
 {\ifodd\value{ALG@line}\relax\ifnum\value{ALG@line}<10\shadeline{4.6pt}\else\shadeline{0pt}\fi\fi\footnotesize \arabic{ALG@line}:}

\usepackage[acronym]{glossaries}
\glsdisablehyper
\newacronym{ml}{ML}{Machine Learning}
\newacronym{mdp}{MDP}{Markov Decision Process}
\newacronym{pomdp}{POMDP}{Partially Observable Markov Decision Process}
\newacronym{rl}{RL}{Reinforcement Learning}
\newacronym{a2c}{A2C}{Advantage Actor Critic}
\newacronym{sac}{SAC}{Soft Actor Critic}
\newacronym{trpo}{TRPO}{Trust Region Policy Optimization}
\newacronym{mbpo}{MBPO}{Model-Based Policy Optimization}
\newacronym{mbcd}{MBCD}{Model-Based Changepoint Detection}
\newacronym{ppo}{PPO}{Proximal Policy Optimization}
\newacronym{a3c}{A3C}{Asynchronous Advantage Actor Critic}
\newacronym{gae}{GAE}{Generalized Advantage Estimation}
\newacronym{dqn}{DQN}{Deep Q Network}
\newacronym{ddqn}{DDQN}{Double Deep Q Network}
\newacronym{dnn}{DNN}{Deep Neural Network}
\newacronym{nn}{NN}{Neural Network}
\newacronym{abr}{ABR}{Adaptive Bit Rate}
\newacronym{bba}{BBA}{Buffer Based Approach}
\newacronym{qoe}{QoE}{Quality of Experience}
\newacronym{cf}{CF}{Catastrophic Forgetting}
\newacronym{ood}{OOD}{Out-of-Distribution}
\newacronym{cpd}{CPD}{Change-Point Detection}
\newacronym{ewc}{EWC}{Elastic Weight Consolidation}
\newacronym{ogd}{OGD}{Orthogonal Gradient Descent}

\theoremstyle{plain}
\newtheorem{theorem}{Theorem}[section]

\newtheorem{lemma}[theorem]{Lemma}

\theoremstyle{definition}

\theoremstyle{remark}

\usepackage{iclr2025_conference}

\iclrfinalcopy 

\author{%
Pouya Hamadanian\\
MIT CSAIL\\
\texttt{pouyah@mit.edu}\\
\And
Arash Nasr-Esfahany\\
MIT CSAIL\\
\texttt{arashne@mit.edu}\\
\And
Malte Schwarzkopf\\
CS Brown University\\
\texttt{malte@cs.brown.edu}\\
\And
Siddhartha Sen\\
Microsoft Research\\
\texttt{sidsen@microsoft.com}\\
\And
Mohammad Alizadeh\\
MIT CSAIL\\
\texttt{alizadeh@mit.edu}\\
}

\title{Online Reinforcement Learning in\\Non-Stationary Context-Driven Environments}

\begin{document}

\maketitle

\begin{abstract}

We study online reinforcement learning (RL) in non-stationary environments, where a time-varying exogenous context process affects the environment dynamics. Online RL is challenging in such environments due to ``catastrophic forgetting'' (CF). The agent tends to forget prior knowledge as it trains on new experiences. Prior approaches to mitigate this issue assume task labels (which are often not available in practice), employ brittle regularization heuristics, or use off-policy methods that suffer from instability and poor performance. 

We present \bigsys (\sys), an online RL approach that combats CF by anchoring policy outputs on old experiences while optimizing the return on current experiences. To perform this anchoring, \sys locally constrains policy optimization using samples from experiences that lie outside of the current context distribution. We evaluate \sys in Mujoco, classic control and computer systems environments with a variety of synthetic and real context traces, and find that it outperforms a variety of baselines in the non-stationary setting, while achieving results on-par with a ``prescient'' agent trained offline across all context traces.

\sys's source code is available at \url{https://github.com/pouyahmdn/LCPO}.

\end{abstract}

\section{Introduction}
\label{sec:intro}
\noindent
{\em --- Those who cannot remember the past are condemned to repeat it. (George Santayana, The Life of Reason, 1905)}

\gls{rl} has seen success in many domains~\citep{mao2017neural, haarnoja2018composable, mao2019learning, marcus2019neo, zhu2020ingredients, haydari2022deeprl}, but real-world deployments have been rare. A major hurdle has been the gap between simulation and reality, where the environment simulators do not match the real-world dynamics. Thus, recent work has turned to applying \gls{rl} in an {\em online} fashion, i.e., continuously training and using an agent in a live environment~\citep{zhang2021loki, gu2021knowledge}.

While online \gls{rl} is difficult in and of itself, it is particularly challenging in {\em non-stationary} environments---also known as continual \gls{rl}~\citep{khetarpal2020towards}---where the characteristics of the environment change over time. A key challenge is \gls{cf}~\citep{MCCLOSKEY1989109}. An agent based on function approximators like \glspl{nn} tends to forget its past knowledge when training sequentially on new non-stationary data. On-policy \gls{rl} algorithms~\citep{sutton_intro_rl} are particularly vulnerable to \gls{cf} in non-stationary environments, since these methods cannot retrain on stale data from prior experiences.

In this paper, we consider problems where the source of the non-stationarity is an observed exogenous {\em context} process that varies over time and exposes the agent to different environment dynamics. Such context-driven environments~\citep{sinclair2023hindsight, mao2018learning, zhang2023robust, Dietterich2018Discovering, pan2022isodream} appear in a variety of applications. Examples include computer systems subject to incoming workloads~\citep{mao2018learning}, locomotion in environments with varying terrains and obstacles~\citep{heess2018emergence}, robots subject to external forces~\citep{pinto2017robust}, and more. In contrast to most prior work~\citep{alegre2021minimum,chandak2020optimizing}, we do not restrict the context process to be discrete, piecewise stationary or Markov.

Broadly speaking, there are three existing approaches to mitigate \gls{cf} in online learning.
One class of techniques is {\em task-based}~\citep{rusu2016progressive, kirkpatrick2017overcoming, schwarz2018progress, farajtabar2019orthogonal,zeng2019continual}. Such works assume explicit task labels that identify the different context distributions which the agent encounters over time. 
Task labels make it easier to prevent the training for one context from affecting  knowledge learned for other contexts. 
In settings where task labels (or boundaries) are not available, a few proposals try to infer the task labels via self-supervised~\citep{nagabandi2019deep} or \gls{cpd} approaches~\citep{Padakandla_2020, alegre2021minimum}. These techniques, however, are brittle when the context processes are difficult to separate and task boundaries are not pronounced~\citep{hamadanian2022demistify}. Our experiments show that erroneous task labels lead to poorly performing agents in such environments (\S\ref{sec:eval}).

A second category of approaches avoids task labels by approximating task-based methods with heuristics~\citep{schwarz2018progress,chaudhry2018riemannian,kaplanis2018continual,woo2022structure}. However, these heuristics are based on brittle assumptions about the nature and cadence of non-stationarity. For example, one approach implicitly assumes each episode is a distinct task, and uses a window of past $N$ episode to regularize learning~\citep{woo2022structure}. These assumptions are rarely met and would likely lead to poor performance in practice, as we observe in our analysis and evaluations (\S\ref{sec:eval} and \S\ref{subsec:base_ewc}, \S\ref{subsec:base_ogd} and \S\ref{subsec:base_bfdqn} in the Appendix).

The third category of approaches employs rehearsal, i.e., learning using past or generated data. 
For example, off-policy learning~\citep{sutton_intro_rl} makes it possible to retrain on past data. These techniques (e.g., Experience Replay~\citep{mnih2013playing}, CLEAR~\citep{rolnick2019experience}, etc.) store prior experience data in a buffer and sample from the buffer randomly to train. Not only does this improve sample complexity, it sidesteps the pitfalls of sequential learning and prevents \gls{cf}~\citep{rolnick2019experience}. However, off-policy methods come at the cost of increased hyper-parameter sensitivity and unstable training~\citep{duan2016benchmarking, gu2016q, haarnoja2018soft}. This brittleness is particularly catastrophic in an online setting, as we also observe in our experiments (\S\ref{sec:eval}). 

We present \sys (\S\ref{sec:motivation}), an on-policy \gls{rl} algorithm that ``anchors'' policy outputs for old contexts while optimizing for the current context. Unlike prior work, \sys does not rely on task labels and only requires an \gls{ood} detector, i.e., a function that recognizes old experiences that occurred in a sufficiently different context than the current one. \sys maintains a bounded buffer of past experiences, similar to off-policy methods~\citep{mnih2013playing}. But as an on-policy approach, \sys does not use stale experiences to optimize the policy. Instead, it uses past data to {\em constrain} the policy optimization on fresh data, such that the agent's behavior does not change in other contexts. 

We evaluate \sys on several environments with real and synthetic contexts (\S\ref{sec:eval}), and show that it outperforms a variety of baselines across mentioned categories in the online learning setting. We also compare against a ``prescient agent'' that is trained offline on the entire context distribution prior to deployment. The prescient agent does not suffer from \gls{cf}. Among all the online methods, \sys is the closest to this idealized baseline. Our ablation results show that \sys is robust to variations in the \gls{ood} detector's thresholds and works well with small experience buffer sizes.

\sys's source code is available online at \url{https://github.com/pouyahmdn/LCPO}.
\section{Preliminaries}
\label{sec:background}

\label{subsec:notation}
\paragraph{Notation.}
We consider online reinforcement learning in a non-stationary context-driven \gls{mdp}, where the context is observed (only up to the current time step $t$) and exogenous. 
Formally, at time step $t$ the environment has state $s_t \in \mathcal{S}$ and context $z_t \in \mathcal{Z}$. The agent takes action $a_t\in \mathcal{A}$ based on the observed state $s_t$ and context $z_t$, $a_t=\pi(s_t, z_t)$, and receives feedback in the form of a scalar reward $r_t = r(s_t, z_t, a_t)$, where $r(\cdot, \cdot, \cdot): \mathcal{S}\times\mathcal{Z}\times\mathcal{A} \rightarrow \mathbb{R}$ is the reward function. The environment's state, the context, and the agent's action determine the next state, $s_{t+1}$, according to a transition kernel, $T(s_{t+1}|s_t, z_t, a_t)$. The context $z_t$ is an independent stochastic process, unaffected by states $s_t$ or actions $a_t$. Finally, $d_0$ defines the distribution over initial states ($s_0$). This model is fully defined by the tuple $\mathcal{M}=(\mathcal{S}, \mathcal{Z}, \mathcal{A}, \{z_t\}_{t=1}^{\infty}, T, d_0, r)$.

\label{subsec:nonstat_rl}
\paragraph{Non-stationary contexts.}
The non-stationary context $\mathbf{z}=\{z_t\}_{t=1}^{\infty}$ impacts the environment dynamics and implies a non-stationary environment. We assume the context process can change arbitrarily: e.g., it can follow a predictable pattern, be i.i.d samples from some distribution, be a discrete process or a multi-dimensional continuous process, experience smooth or dramatic shifts, be piecewise stationary, or include any mixture of the above. We have no prior knowledge of the context process, the environment dynamics, or access to an offline simulator. Examples of (observed) context processes include market demand in a supply chain system, incoming request workloads in virtual machine allocation problem, customer distributions in airline revenue management~\citep{sinclair2023hindsight}, traffic information in vehicular networks~\citep{wu2017flow}, terrain profiles in a locomotion task~\citep{heess2017emergence}, network traffic for video streaming~\citep{mao2020realworld} and congestion control~\citep{winstein2013tcp}, etc.

\label{subsec:goal}
\paragraph{Goal.}
We seek good long-term performance. Formally, for a given policy $\pi: \mathcal{S} \times \mathcal{Z} \rightarrow \mathcal{A}$ and context process $\mathbf{z}=\{z_t\}_{t=1}^{\infty}$ we define the lifelong return as $J(\pi, \mathbf{z})=\lim_{t\rightarrow\infty} \sum_{i=1}^{t}\frac{r_i}{t}$ for an infinite horizon \gls{mdp}. Similarly, for finite horizon \glspl{mdp} of length $H$, where episode $i$ is subject to context traces $\mathbf{z}_i=(z_{H.i}, z_{H.i+1},...,z_{H.(i+1)-1})$ and has an episodic return of $R_i=\sum_{t=1}^{H}r_t^{(i)}$, we define the lifelong return as $J(\pi, \mathbf{z})=\lim_{t\rightarrow\infty} \sum_{i=1}^{t}\frac{R_i}{t}$. For a policy sequence $\mathbf{\Pi}=\{\pi_t\}_{t=1}^{\infty}$, e.g., the sequence of policies resulting from a continual \gls{rl} algorithm, we can define the lifelong return $J(\mathbf{\Pi}, \mathbf{z})$ similarly.
 
\label{subsec:online_rl}
\paragraph{Online \gls{rl}.}
In most \gls{rl} settings, a policy is trained in a separate training phase. During testing, the policy is fixed and does not change. By contrast, online learning starts with the test phase, and the policy must reach and maintain optimality within this test phase. An important constraint in the online setting is that the agent gets to experience each interaction only once. There is no way to revisit past interactions and try a different action in the same context. This is a key distinction with training in a separate offline phase, such as in meta-learning~\citep{alshedivat2018continuous}, where the agent can explore the same conditions many times.

Note that the policy that maximizes lifelong return $\pi^*=arg\max_{\pi} J(\pi,\mathbf{z})$ has to be \emph{prescient}, i.e., it needs to have upfront knowledge of the context process $\mathbf{z}$. Since an online agent is causal and has only observed context values up to the current time step $t$, it can never perform as well as this prescient policy. Therefore, in general online \gls{rl} agents will have a gap with prescient policies in terms of lifelong return. In certain special cases the online \gls{rl} can asymptotically reach the prescient policy, e.g., when the context process is Markovian the entire context-driven \gls{mdp} collapses to a standard \gls{mdp} with a state $\tilde{s}_t=<s_t, z_t>$. However, we do not intend to limit the context process in any way, and our aim it to minimize the gap between the online and prescient agents for arbitrary context processes.

\section{Related Work}
\label{sec:related_works}

\paragraph{Non-stationary \gls{rl}.}
Non-stationary \gls{rl} is a family of sub-problems, such as \gls{cf}, latent context inference, meta-learning, etc~\citep{khetarpal2020towards}. In this work we focus on \gls{cf}, and highlight the differences of \gls{cf} with other well-known non-stationary \gls{rl} problems below. Then, we will explore related work for \gls{cf} in \gls{ml} and \gls{rl}. We highlight other lines of work in \S\ref{sec:related_works_cont} in the Appendix.

\paragraph{Latent Context Inference.}
These works consider a context-driven \gls{mdp} where the context $z_t$ is unobserved. The goal is to infer an estimated context $\hat{z_t}$ from other signals, such as transition functions, reward functions, etc~\citep{hallak2015contextualmarkovdecisionprocesses,zintgraf2019fast,xie2020deep,caccia2020online,lee2020contextaware,he2020task,poiani2021meta,chen2022adaptive,huang2022adarl,feng2022factored,ren2022hdpcmdp,woo2022structure,bing2022metareinforcement, luo2022adapt,lee2023tempo}. Once inferred, a traditional \gls{rl} algorithm such as \gls{sac} learns a policy $\pi(\cdot|s_t, \hat{z_t})$ from the state and inferred context, and is typically compared to an `upper-bound policy' that observes the true context $\pi(\cdot|s_t, z_t)$. These works aim to recover the unobserved context, while we focus on \gls{cf} after observing the true/recovered context. In fact, the `upper-bound' policies in these works are baselines we compare to in \S\ref{sec:eval}. Combining \sys with this line of work to solve \gls{cf} in environments with latent context is an interesting future work.

\paragraph{Catastrophic Forgetting.}
Three general techniques exist for mitigating \gls{cf} in \gls{ml}~\citep{PARISI201954}; (1)~regularizing the optimization to avoid memory loss during sequential training~\citep{kirkpatrick2017overcoming,zenke2017continual,farajtabar2019orthogonal,lopezpaz2022gradient}; (2)~training separate parameters per task, and expanding/shrinking parameters as necessary~\citep{rusu2016progressive,shmelkov2017incremental,li2019learn}; (3)~rehearsal, i.e., retraining on original data or generative batches~\citep{shin2017continual,isele2018selective,Atkinson_2021}; or combinations of these techniques~\citep{schwarz2018progress,aljundi2019taskfree}.

Regularization techniques such as \gls{ewc} and \gls{ogd} require task labels. Approximations have been proposed for problems without task labels~\citep{schwarz2018progress} or boundaries~\citep{woo2022structure}. \citet{kaplanis2018continual} use biologically inspired models of brain synapses to regularize networks.

Another class of approaches aims to infer task labels, by learning the transition dynamics of the \gls{mdp}, and detecting a new environment when a surprising sample is observed with respect to the learned model~\citep{doya2002multiple,rlcd,Padakandla_2020,alegre2021minimum}. The inferred labels are often used to train separate policies/models to mitigate \gls{cf}. These methods are effective when \gls{mdp} transitions are abrupt and have well-defined boundaries, but are brittle and perform poorly in realistic environments with noisy and hard-to-distinguish non-stationarities~\citep{hamadanian2022demistify}.

For rehearsal, we can use learned models of the \gls{mdp} to replay past experiences~\citep{xu2020taskagnostic,lee2020contextaware,pong2020temporal,huang2021continual,janner2021trust}.
An alternative type of rehearsal is off-policy training~\citep{haarnoja2018soft,vanhasselt2016learning} (e.g., Experience Replay~\citep{mnih2013playing}), which can train on stale data, naturally circumvent sequential learning and avoid \gls{cf}. However, off-policy \gls{rl} is empirically unstable and sensitive to hyperparameters due to bootstrapping and function approximation~\citep{sutton_intro_rl}, and is often outperformed by on-policy algorithms in online settings~\citep{duan2016benchmarking, gu2016q, haarnoja2018soft}. CLEAR~\citep{rolnick2019experience} is an off-policy \gls{rl} algorithm explicitly designed to overcome \gls{cf} with fast adaptations. Similarly, PT-DQN \citep{anand2023predictioncontrolcontinualreinforcement} learns a permanent Q-network to remember past tasks while learning a transient Q-network for fast adaptation.

\paragraph{Constrained Optimization.} 
\sys's constrained optimization formulation is structurally similar to \gls{trpo}~\citep{schulman2015trust}, despite our different problem setting.

\section{Locally-Constrained Policy Optimization}

Our goal is to learn a policy $\pi(\cdot, \cdot)$ that takes action $a_t \sim \pi(s_t, z_t)$, in a context-driven \gls{mdp} characterized by an exogenous non-stationary context process. 

\subsection{Illustrative Example}
\label{sec:motivation}
\begin{figure}[t]
    \centering
    \hspace{0.02\linewidth}
    \begin{subfigure}{0.2\linewidth}
        \includegraphics[page=8, width=\linewidth]{figures/LCPO_figures-crop.pdf}
        \caption{}
        \label{fig:grid_vis_no_trap}
    \end{subfigure}
    \hspace{0.02\linewidth}
    \begin{subfigure}{0.2\linewidth}
        \includegraphics[page=9, width=\linewidth]{figures/LCPO_figures-crop.pdf}
        \caption{}
        \label{fig:grid_vis_trap}
    \end{subfigure}
    \hspace{0.03\linewidth}
    \begin{subfigure}{0.48\linewidth}
        \centering
        \includegraphics[width=\linewidth]{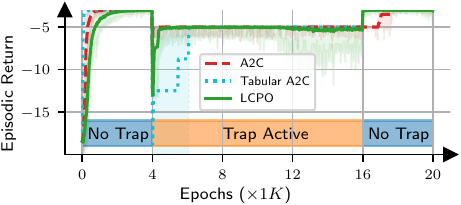}
        \caption{}
        \label{fig:grid_rew}
    \end{subfigure}
    \begin{subfigure}{0.48\linewidth}
        \centering
        \includegraphics[width=\linewidth]{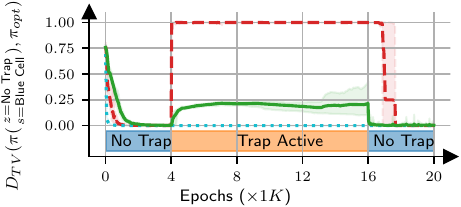}
        \caption{}
        \label{fig:grid_out_no_trap}
    \end{subfigure}
    \hspace{0.02\linewidth}
    \begin{subfigure}{0.48\linewidth}
        \centering
        \includegraphics[width=\linewidth]{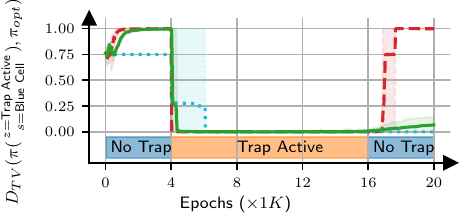}
        \caption{}
        \label{fig:grid_out_trap_active}
    \end{subfigure}
    \caption{A 3x3 grid-world problem with two modes and the optimal path visualized in blue. \textbf{(a)} In the `No Trap' mode, the center square is safe to pass through. \textbf{(b)} In the `Trap` mode, the agent must avoid the trap with a longer path. \textbf{(c)} Episodic return across time in the grid environment. \textbf{(d and e)} Total variation distance between learned and optimal policy outputs for the \textbf{(d)} `No Trap' mode, and the \textbf{(e)} `Trap Active' mode at the blue cell (lower is better). Tabular A2C and LCPO remember the optimal decision for either context during shaded regions and instantly attain optimal returns when the environment switches.}
    \label{fig:grid_vis}
\end{figure}

Consider a simple environment with a discrete context. In this grid-world problem depicted in \Cref{fig:grid_vis}, the agent can move in 4 directions in a 2D grid, and incurs a base negative reward of $-1$ per step until it reaches the terminal exit state (no penalty in the last step) or fails to reach the exit within $20$ steps. The grid can be in two modes; 1) `No Trap' mode, where the center cell is empty, and 2) `Trap Active' mode, where walking into the center cell incurs a reward of $-10$. When in `No Trap' mode, the optimal path passes through the center cell, and the best episodic return is $-3$. In the `Trap Active' mode, the center cell's penalty forces the optimal path to go left at the blue cell for an optimal episodic return of $-5$. This environment mode is our discrete context and the source of non-stationarity in this simple example. The agent observes its current location and the context, i.e., whether the trap is on the grid ($z_t=1$) or not ($z_t=0$) in every episode (beginning from the start square).

\paragraph{\gls{a2c}.} We use the \gls{a2c} algorithm to train a policy for this environment, while its context changes every so often. \Cref{fig:grid_rew} depicts the episodic return across time and \Cref{fig:grid_out_no_trap,fig:grid_out_trap_active} depict the total variation distance between the optimal and learned policy when the policy input is `No Trap' mode (\Cref{fig:grid_out_no_trap}) or `Trap Active' mode (\Cref{fig:grid_out_trap_active}). This distance represents how close the learned policy is to the optimal in either context. The agent initially attains optimality for the `No Trap' mode, but once the context changes at epoch 4K it immediately forgets it. Note that during epochs 4K-16K, the \gls{a2c} agent is only trained on samples from the `Trap Active' mode, and its output for the `No Trap' mode is drifting. When the context changes back to the `No Trap' mode at epoch 16K, the agent behaves sub-optimally (epochs 16K-18K) before relearning. \Cref{fig:grid_out_trap_active} shows that \gls{a2c} also forgets the optimal `Trap Active' policy during the final 4K epochs.

\paragraph{Key Insight.} Since the policy observes the current context $z_t$, it should be able to distinguish between different environment modes. Therefore, if the agent could surgically modify its policy on the current state-context pairs $\pi(\cdot|s_t, z_t)$ and leave outputs for other state-context pairs $\pi(\cdot|s_t, z'\neq z_t)$ unchanged, it would eventually learn a good policy for all contexts. In fact, tabular RL achieves this trivially in this finite discrete state-context space. To illustrate, we apply a tabular version of \gls{a2c}: i.e., the policy and value networks are replaced with tables with separate rows for each state-context pair (18 total rows). \Cref{fig:grid_out_no_trap,fig:grid_out_trap_active} demonstrate that the tabular RL policy for each context remains unchanged when it does not actively interact with that context. This is because when an experience is used to update the table, it only updates the row pertaining to its own state and context, and does not change rows belonging to other contexts. Under sufficient conditions, tabular RL can provably converge to the optimal policy for such environments. Due to space constraints, we state the theorem and its proof in \S\ref{sec:theory} in the Appendix.

Can we achieve a similar behavior with neural network function approximators? In general, updating a neural network (say, a policy network) for certain state-context pairs will change the output for all state-context pairs, leading to \gls{cf}. But if we could somehow ``anchor'' the output of the neural network on distinct prior state-context pairs (analogous to the cells in the tabular setting) while we update the relevant state-context pairs, then the neural network would not ``forget''.

\paragraph{\sys.} Achieving the aforementioned anchoring does not require task labels. We only need to know if two contexts $z_i$ and $z_j$ are {\em different}. In particular, let the batch of recent environment interactions $<s_t, z_t, a_t, r_t>$ be $B_{r}$ and let all previous interactions (from possibly different contexts) be $B_{a}$. Suppose we have a difference detector $W(B_a, B_r)$ that can be used to sample experiences from $B_a$ that are not from the same distribution as the samples in the recent batch $B_r$, i.e., the difference detector provides out-of-distribution (OOD) samples with respect to $B_r$. Then, when optimizing the policy for the current batch $B_{r}$, we can constrain the policy's output on experiences sampled via $W(B_a, B_r)$ to not change (see \S\ref{sec:method} for details). We name this approach \bigsys (\sys). The result for \sys is presented in \Cref{fig:grid_out_no_trap,fig:grid_out_trap_active}. While it does not retain its policy as perfectly as tabular \gls{a2c}, it does sufficiently well to recover near instantaneously upon the second switch at epoch 16K.

\begin{figure}[t]
    \centering
    \begin{subfigure}{0.45\linewidth}
        \centering
        \includegraphics[width=\linewidth]{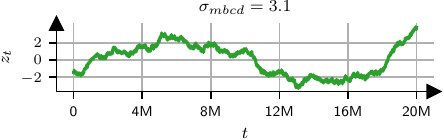}
    \end{subfigure}
    \hspace{0.15in}
    \begin{subfigure}{0.45\linewidth}
        \centering
        \includegraphics[width=\linewidth]{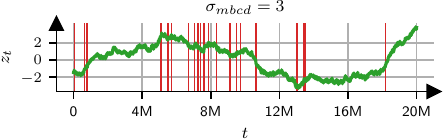}
    \end{subfigure}
    \caption{A sample context process $z_t$, and detected change-points at two thresholds. Teasing meaningful task boundaries is difficult for this process, but defining an \gls{ood} metric is intuitive.}
    \label{fig:ood_vs_cpd}
\end{figure}

\paragraph{Change-Point Detection (CPD) vs. OOD Detection.} \gls{cpd} (and task labeling in general) requires stronger assumptions than \gls{ood} detection. The context process has to be piecewise stationary to infer task labels and context changes must happen infrequently to be detectable. Furthermore, online \gls{cpd} is sensitive to outliers. In contrast, \gls{ood} is akin to defining a distance metric on the context process and can be well-defined on any context process.
Consider the context process shown in \Cref{fig:ood_vs_cpd}. We run this context process through a \gls{cpd} algorithm~\citep{alegre2021minimum} for two different sensitivity factors $\sigma_{mbcd}$, and represent each detected change-point with a red vertical line. A slight increase in sensitivity leads to 34 detected change-points, and these change-points are also not reasonable. There is no obvious way to assign task labels for this smooth process and there aren't clearly separated segments that can be defined as tasks. However, an intuitive \gls{ood} detection method is testing if the distance between $z_i$ and $z_j$ is larger than some threshold, i.e., $|z_i-z_j|>1$.
Altogether, \gls{ood} is considerably easier in practice compared to \gls{cpd}. Note that although the grid-world example~--~and discrete context environments in general~--~is a good fit for \gls{cpd}, this environment was purposefully simple to explain the insight behind \sys.

\subsection{Methodology}
\label{sec:method}

Consider a parameterized policy $\pi_\theta$ with parameters $\theta$. Our task is to choose a direction for changing $\theta$ such that it improves the expected return on the most recent batch of experiences $B_r$, while the policy is `anchored' on prior samples with sufficiently distinct context distributions, $W(B_a, B_r)$.

\begin{algorithm}[ht]
\caption{\sys Training}
\label{algo:train}
\begin{algorithmic}[1]
\State initialize parameter vectors $\theta_0$, empty buffer $B_a$
\For{each iteration}
    \State $B_r \gets$ Sample a mini-batch of new interactions
    \State $S_c \gets W(B_a, B_r)$
    \State $v \gets \nabla_{\theta}\mathcal{L}_{tot}(\theta; B_r)\vert_{\theta_0}$
    \If {$S_c$ is not empty}
        \State $g(x) := \nabla_{\theta}(x^T\nabla_{\theta}\mathcal{D}_{KL}(\theta_{old}, \theta; S_{c})\vert_{\theta_0})\vert_{\theta_0}$
        \State $v_{c} \gets conjgrad(v, g(\cdot))$
        \While {$\theta_{old} + v_{c}$ violates constraints}
            \State $v_{c} \gets v_{c}/2$
        \EndWhile
        \State $\theta_0 \gets \theta_0 + v_{c}$
    \Else
        \State $\theta_0 \gets \theta_0 + v$
    \EndIf
    \State $B_a \gets B_a + B_r$
    
\EndFor
\end{algorithmic}
\end{algorithm}

In supervised learning, this anchoring is straightforward to perform, e.g., by adding a regularization loss that directs the neural network to output the ground truth labels for \gls{ood} samples~\citep{caruana1997multitask}. In the case of an RL policy, however, we do not know the ground truth (optimal actions) for anchoring the policy output. Moreover, using the actions we took in prior contexts as the ground truth is not possible, since the policy may have not converged at those times. Anchoring to those actions may cause the policy to relearn suboptimal actions from an earlier period in training. 
To avoid these problems, \sys solves a constrained optimization problem that forces the policy to not change for \gls{ood} samples. Formally, we consider the following optimization problem:
\begin{equation}
\label{eqn:sys_goal}
    \centering
    \begin{aligned}
        \min_{\theta}~~ \mathcal{L}_{tot}(\theta; B_r) \triangleq \mathcal{L}_{PG}(\theta; B_r)& + \mathcal{L}_{e}(\theta; B_r)  \\
        s.t.~~ D_{KL}(\theta_0, \theta; W(B_a, B_r&)) \le c_{anchor} \\
    \end{aligned}
\end{equation}
We use the standard definition of policy gradient loss, that optimizes a policy to maximize returns~\citep{schulman2018highdimensional, mnih2016asynchronous, sutton_intro_rl}:
\begin{equation}
    \centering
    \begin{aligned}
        \mathcal{L}_{PG}(\theta; B_r)= \mathbb{E}_{r_t \sim B_r}\left[ \sum_{t=0}^{H} -\gamma^t r_t \right]
    \end{aligned}
\end{equation}

We use automatic entropy regularization~\citep{haarnoja2018softapps}, to react to and explore in response to novel contexts. The learnable parameter $\theta_{e}$ is adapted such that the entropy coefficient is $e^{\theta_{e}}$, and the entropy remains close to a target entropy $\bar{\mathcal{H}}$. This worked well in our experiments but \sys could use any exploration method designed for non-stationary context-driven environments.
\begin{equation}
    \centering
    \begin{aligned}
        \mathcal{L}_{e}(\theta; B_r)=
        e^{\theta_{e}}
\mathbb{E}_{s_{t},z_{t} \sim B_r, a_{t}\sim\pi} \left[\log\pi(a_t|s_t,z_t)\right] \\
    \end{aligned}
\end{equation}

We use KL-divergence as a measure of policy change, and for simplicity we use $D_{KL}(\theta_0, \theta; W(B_a, B_r))$ as a shorthand for $\mathbb{E}_{s, z \sim W(B_a, B_r)}[D_{KL}(\pi_{\theta_0}(s, z) || \pi_\theta(s, z))]$. Here, $\theta_0$ denotes the current policy parameters, and we are solving the optimization over $\theta$ to determine the new policy parameters.

\paragraph{Buffer Management.} To avoid storing all interactions in $B_a$, we use reservoir sampling~\citep{vitter1985random}; we randomly replace old interactions with new ones with probability $\frac{n_{b}}{n_{s}}$, where $n_b$ is the buffer size and $n_s$ is the total interactions thus far. Reservoir sampling guarantees that the interactions in the buffer are a uniformly random subset of the full set of interactions. For a pseudo-code see \S\ref{sec:lcpo_reservoir} in the Appendix.

\paragraph{Difference detector.} To realize $W(B_a, B_r)$, we treat it as an \gls{ood} detection task. A variety of methods can be used in practice (\S\ref{sec:eval}), e.g., we can compute the Mahalanobis distance~\citep{mahalanobis2018generalized}~---~the normalized distance of each experience's context with respect to the average context in $B_r$~---~and deem any distance above a certain threshold to be \gls{ood}. To avoid a high computational overhead when sampling from $W(B_a, B_r)$, we sample a larger batch from $B_a$, and keep the state-context pairs that are \gls{ood} with respect to $B_r$. If not enough different samples exist, we do not apply the constraint for that update. For a  pseudo-code and further implementation details about the \gls{ood} detector, see \S\ref{sec:lcpo_ood_dist} and \S\ref{sec:lcpo_ood_sample}.

\paragraph{Solving the constrained optimization.} To solve this constrained optimization, we approximate the optimization goal and constraint, and calculate a search direction accordingly (pseudocode in \Cref{algo:train}). Our problem is structurally similar to \gls{trpo}~\citep{schulman2015trust}, though the constraint is quite different. Similar to \gls{trpo}, we model the optimization goal with a first-order approximation, i.e., $\mathcal{L}_{tot}(\theta; \cdot) = \mathcal{L}_0 + (\theta-\theta_0)^T\nabla_{\theta}\mathcal{L}_{tot}(\theta; \cdot)\vert_{\theta_0}$, and the constraint with a second order approximation $D_{KL}(\theta_0, \theta; \cdot)=(\theta-\theta_0)^T\nabla^2_{\theta}D_{KL}(\theta_0, \theta; \cdot)\vert_{\theta_0}(\theta-\theta_0)$. The optimization problem can therefore be written as
\begin{equation}
    \begin{aligned}
        \min_{\theta}~~ (\theta-\theta_0)^T &v \\
        s.t.~~ (\theta-\theta_0)^T A(\theta-&\theta_0) \le c_{anchor} \\
    \end{aligned}
\end{equation}
where $A_{ij}=\frac{\partial}{\partial \theta_i}{\frac{\partial}{\partial \theta_j}} D_{KL}(\theta_0, \theta; W(B_a, B_r))$, and $v=\nabla_{\theta}\mathcal{L}_{tot}(\theta; \cdot)\vert_{\theta_0}$. This optimization problem can be solved using the conjugate gradient method followed by a line search~\citep{schulman2015trust,achiam2017constrained}.

\paragraph{Bounding policy change.} The above formulation does not bound policy change on the current context, which could destabilize learning. We could add a second constraint, i.e., \gls{trpo}'s constraint, $D_{KL}(\theta_0, \theta; B_r) \le c_{recent}$ (note that this constraint is different from that in \Cref{eqn:sys_goal}, as the samples come from $B_r$ instead of $W(B_a, B_r)$). However, having two second order constraints is computationally expensive.
Instead, we guarantee the \gls{trpo} constraint in the line search phase (lines 9--10 in \Cref{algo:train}), where we repeatedly decrease the gradient update size until both constraints are met. 
\section{Evaluation}
\label{sec:eval}

We evaluate \sys across six environments: four from Gymnasium Mujoco~\citep{towers2023gymnasium}, one from Gymnasium Classic Control~\citep{towers2023gymnasium}, and a straggler mitigation task from computer systems~\citep{mao2019towards,hamadanian2022demistify}. These environments are subject to synthetic or real context processes that affect their dynamics. Our experiments aim to answer the following questions: \textbf{(1)} How does \sys compare to baselines, and can it perform as well as the pre-trained prescient policies (\S\ref{sec:all_eval})? 
\textbf{(2)} How does the accuracy of the \gls{ood} sampler $W(\cdot, \cdot)$ affect \sys (\S\ref{sec:ood_agg_eval})?
\textbf{(3)} How does the maximum buffer size $n_b=|B_a|$ affect \sys (\S\ref{sec:ood_size_eval})?
We include further ablations of \sys and baselines in Appendices \S\ref{sec:lcpo_p}, \S\ref{subsec:base_mbpo} and \S\ref{subsec:base_ewc}

\paragraph{Baselines.} We consider the following approaches for comparison:
\textbf{Regularization-based:} (1) Online \gls{ewc}~\citep{kirkpatrick2017overcoming,chaudhry2018riemannian, schwarz2018progress}, (2) Sliding \gls{ogd}~\citep{farajtabar2019orthogonal,woo2022structure} and (3) Benna-Fusi DQN (BFQDN)~\citep{kaplanis2018continual}, \textbf{Task Inference:} (4) \gls{mbcd}~\citep{alegre2021minimum}, \textbf{Rehearsal:} (5) \gls{mbpo}~\citep{janner2021trust}, (6) CLEAR~\citep{rolnick2019experience}, (7) PT-DQN~\citep{anand2023predictioncontrolcontinualreinforcement}, (8) \gls{sac}~\citep{haarnoja2018soft} and (9) \gls{ddqn}~\citep{doubledqn} \textbf{On-policy RL:} (10) \gls{a2c}~\citep{mnih2016asynchronous} and (11) \gls{trpo} (single-path)~\citep{schulman2015trust}, both using \gls{gae}~\citep{schulman2018highdimensional}, \textbf{Prescient RL:} (12) as described in \S\ref{sec:background}, the best of policies trained with \gls{a2c}~\citep{mnih2016asynchronous}, \gls{trpo} (single-path)~\citep{schulman2015trust}, \gls{ddqn}~\citep{doubledqn} and \gls{sac}~\citep{haarnoja2018soft}. For more details about these baselines, refer to \S\ref{sec:baselines} in the Appendix

\paragraph{Experiment Setup.} We use 25 random seeds for gymnasium (5 for slower schemes) and 10 random seeds for the straggler mitigation experiments, and use the same hyperparameters for \sys in all environments and contexts. Gym environments were modified to accept discrete action space policies, as even prescient policies struggled to learn stable continuous space policies in the presence of contexts (See \S\ref{sec:pend_discretize}). Hyperparameters and neural network structures are noted in Appendices \S\ref{sec:pend_train_params} and \S\ref{sec:lbalance_train_params}. These experiments were conducted on a machine with 2 AMD EPYC 7763 CPUs (256 logical cores) and 512 GiB of RAM. With 32 concurrent runs, experiments finished in $\sim$1152 hours. This figure does not include runtime devoted to tuning the baselines.

\paragraph{Environment and Contexts.} We consider six environments: Modified versions of Pendulum-v1 from the classic control environments, InvertedPendulum-v4, InvertedDoublePendulum-v4, Hopper-v4 and Reacher-v4 from the Mujoco environments~\citep{towers2023gymnasium}, and a straggler mitigation environment~\citep{hamadanian2022demistify}. In the gym environments, the context is an exogenous ``wind'' process that creates external force on joints and affects movements. We append the external wind vectors from the last 3 time-steps to the observation, since the agent cannot observe the external context that is going to be applied in the next step, and a history of prior steps helps with the prediction. We create 4 synthetic context sequences with the Ornstein–Uhlenbeck process~\citep{uhlorn}, piecewise Gaussian models, or hand-crafted signals with additive noise. These context processes cover smooth, sudden, stochastic, and predictable transitions at short horizons. All context traces are visualized in \Cref{fig:windypend_traces} in the Appendix. Context traces 1 and 2 are 20 million, and context traces 3 and 4 are 8 million steps long. All baselines were allowed a `warm-up' period of 6 million time steps, and episodes were truncated at 200 steps. For the straggler mitigation environments, we use workloads provided by the authors in~\citep{hamadanian2022demistify}, that are from a production web framework cluster at Microsoft, collected from a single day in February 2018. These workloads are visualized in \Cref{fig:lbalance_traces} in the appendix.

\paragraph{\gls{ood} detection.} We set the buffer size $n_b$ to 1\% of all samples, which is $n_b\le200K$. To sample \gls{ood} state-context pairs $W(B_a, B_r)$, we use distance metrics and thresholds. For gym environments where the context is a wind process, we use L2 distance, i.e. if $\overline{w_r}=\mathbb{E}_{w \sim B_r}[w]$ is the average wind vector observed in the recent batch $B_r$, we sample a minibatch of states in $B_a$ where $W(B_a, B_r)=\{w_i|\forall w_i \in B_a: \lVert w_i - \overline{w_r} \rVert_2 > \sigma\}$. There exist domain-specific models for workload distributions in the straggler mitigation environment, but we used Mahalanobis distance as it is a well-accepted and general approach for outlier detection in prior work~\citep{lee2018simple, podolskiy2021revisiting}. Concretely, we fit a Gaussian multivariate model to the recent batch $B_r$, and report a minibatch of states in $B_a$ with a Mahalanobis distance further than $\sigma$ from this distribution (see \S\ref{sec:lcpo_ood_dist} in the Appendix for more details).

\subsection{Results}
\label{sec:all_eval}

\begin{figure}[t!]
    \centering
    \begin{subfigure}{0.49\linewidth}
        \includegraphics[width=\linewidth]{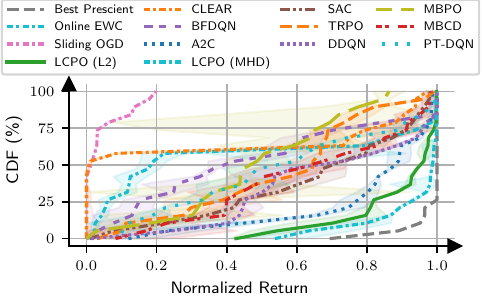}
        \caption{}
        \label{fig:windy_cdf_returns}
    \end{subfigure}
    \begin{subfigure}{0.49\linewidth}
        \includegraphics[width=\linewidth]{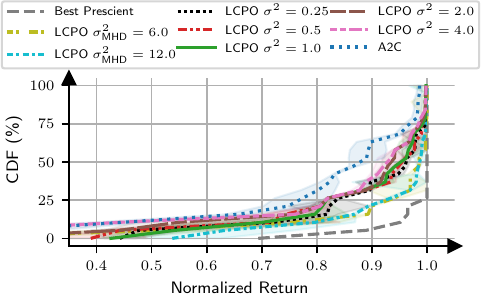}
        \caption{}
        \label{fig:windy_cdf_ablation}
    \end{subfigure}
    \caption{
    CDF of normalized lifelong returns, where 0/1 denote the lowest/highest returns among agents. Shaded regions denote 95\% confidence intervals. \textbf{(a)} \sys outperforms all online agents, and remains the closest to prescient policies. \textbf{(b)} \sys is affected by the \gls{ood} threshold $\sigma$, but still outperforms baselines.}
\end{figure}

To evaluate across different gymnasium environments and traces, we score agents with {\em Normalized Return}, i.e., for each environment and context process we report a scaled score function where 0 and 1 are the minimum and maximum lifelong return across all agents. We prefer agents with higher scores over different environments and traces. \Cref{fig:windy_cdf_returns} provides a summary of all gymnasium experiments (full details in \Cref{table:gym_all_p1,table:gym_all_p2} in the Appendix). \sys maintains a lead over baselines, is close to the best-performing prescient policy, all while learning online and sequentially. We present a detailed analysis of baselines' performance in \S\ref{sec:baselines}, and we summarize these findings below. We also report the wallclock time for each scheme in \S\ref{sec:gym_runtime} in the appendix; \sys is $\sim1.5\times$ as demanding as \gls{a2c}.

Online \gls{ewc} and Sliding \gls{ogd} employ heuristics to circumvent the necessity of task labels in the original techniques. Conceptually, they implicitly assume past episodes are separate ``tasks''. Empirically these heuristics are not successful at solving \gls{cf}. As for BFDQN, \citet{kaplanis2018continual} note that while the architecture was successful in simple environments, it failed with more complex and challenging ones. In our experience, this architecture did not provide any benefits compared to vanilla \gls{ddqn}.

\gls{mbcd} struggles to tease out meaningful task boundaries. In some experiments, it launches anywhere between 3 to 7 policies just by changing the random seed. This observation is in line with \gls{mbcd}'s sensitivity in \S\ref{sec:motivation}, and prior work~\citep{hamadanian2022demistify}.

\gls{mbpo} performed poorly, even though we verified that its learned model of the environment is highly accurate. CLEAR~\citep{rolnick2019experience} and PT-DQN~\citep{anand2023predictioncontrolcontinualreinforcement} are highly hyperparameter sensitive due to how they address catastrophic forgetting. While we tuned both extensively for the Pendulum-v1 environment, as we did for all baselines, they fail catastrophically in other environments. \gls{sac} and \gls{dqn} struggle to outperform \gls{a2c} in the online case. This falls in line with prior observations~\citep{hamadanian2022demistify} and can be attributed to the instability and hyperparameter sensitivity of off-policy \gls{rl}~\citep{duan2016benchmarking, gu2016q, haarnoja2018soft} and the quick adaptation that a fully online algorithm such as \gls{a2c} provides~\citep{sutton2007tracking}. In fact, despite not having been designed for non-stationary \gls{rl}, \gls{a2c} is the most successful baseline.

\begin{table*}[h!]
\caption{Tail latency (negative reward) and 95th percentile confidence ranges for different algorithms and contexts in the straggler mitigation environment.}
\label{table:lbalance_simple}
\scriptsize
\centering
\begin{tabular}{@{}l c c c c c c c c c c c@{}}
\toprule
 & \multicolumn{10}{c}{\textbf{Online Learning}} &\\
\cmidrule{2-11}

& LCPO & LCPO & LCPO & \multirow{2}{*}{MBCD} & \multirow{2}{*}{MBPO} & Online & \multirow{2}{*}{A2C} & \multirow{2}{*}{TRPO} & \multirow{2}{*}{DDQN} & \multirow{2}{*}{SAC} & Best \\
& Agg & Med & Cons & & & EWC & & & & & Prescient\\
\midrule
\multirow{2}{*}{Workload 1} & 1070 & 1076 & 1048 & 1701 & 2531 & 2711 & 1716 & 3154 & 1701 & 1854 & 984 \\
 & $\pm$10 & $\pm$16 & $\pm$7 & $\pm$112 & $\pm$197 & $\pm$232 & $\pm$710 & $\pm$464 & $\pm$47 & $\pm$245 & (TRPO)\\
\midrule
\multirow{2}{*}{Workload 2} & 589 & 617 & 586 & 678 & 891 & 724 & 604 & 864 & 633 & 644 & 509  \\
 & $\pm$43 & $\pm$62 & $\pm$27 & $\pm$38 & $\pm$54 & $\pm$22 & $\pm$109 & $\pm$105 & $\pm$7 & $\pm$27 &  (A2C)\\

\bottomrule

\end{tabular}
\end{table*}

For the straggler mitigation environment, \Cref{table:lbalance_simple} presents the latency metric (negative reward) over two workloads. Recall that this environment uses real-world traces from a production cloud network. The overall trends are similar to the gymnasium experiments, with LCPO outperforming all other baselines. This table includes three variants of \sys, that will be discussed further in \S\ref{sec:ood_agg_eval}.   

\subsection{Sensitivity to OOD metric}
\label{sec:ood_agg_eval}

\sys applies a constraint to \gls{ood} state-context pairs, as dictated by the \gls{ood} sampler $W(B_a, B_r)$. We vary the \gls{ood} threshold $\sigma$---which the \gls{ood} method uses in sampling---and monitor the normalized return for the gym environments in \Cref{fig:windy_cdf_ablation} and the straggler mitigation environments in \Cref{table:lbalance_simple}. In the gym environments, a lower value for $\sigma$ yields tighter margin of difference before a sample is deemed \gls{ood}. \sys is affected by $\sigma$, with the smallest threshold $\sigma^2=0.25$ performing the best. However, \sys still maintains a lead over the \gls{a2c} baseline across $\sigma$ variations. We also experiment with a handicapped \gls{ood} metric that observes a state-context vector $x_t=<s_t, z_t>$ \emph{without the ability to separate state and context.} We use the Mahalanobis distance \gls{ood} metric~\citep{mahalanobis2018generalized} at several thresholds $\sigma^2_{\text{MHD}}$ for this experiment. Despite the handicap, the \sys+Mahalanobis surpasses the \sys+L2 agent that we have used in this evaluation. This is not surprising, as the L2 distance is less robust than Mahalanobis distance, but easier to interpret. In the straggler mitigation environment \sys Agg, \sys Med and \sys Cons use $\sigma=5$, $6$ and $7$, and a higher value for $\sigma$ yields more conservative \gls{ood} samples (i.e., fewer samples are detected as \gls{ood}). The difference between these three is significant: The model in \sys Agg allows for $26.7\times$ more samples to be considered \gls{ood} compared to \sys Cons. \Cref{table:lbalance_simple} provides the normalized return for \sys with varying thresholds, along with baselines. Notably, all variations of \sys achieve similar results.

\subsection{Sensitivity to Buffer Size}
\label{sec:ood_size_eval}

\begin{wrapfigure}{r}{0.5\textwidth}
    \vskip-1.2\baselineskip
    \centering
    \includegraphics[width=0.49\textwidth]{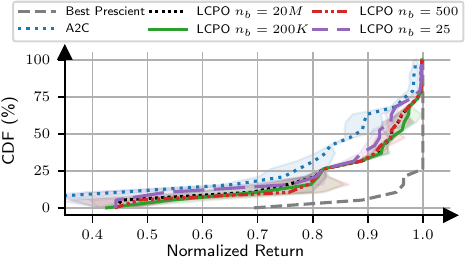}
    \caption{
    CDF of normalized returns of LCPO in gym environments with various buffer sizes. Shaded regions denote 95\% confidence intervals. \sys loses performance with $n_b<500$.}
    \label{fig:windy_buffer_size}
\end{wrapfigure}
\sys uses Reservoir sampling~\citep{vitter1985random} to maintain a limited number of samples $n_b$. 
We evaluate how sensitive \sys is to the buffer size in \Cref{fig:windy_buffer_size} (full results in \Cref{table:gym_buffer_all} in the Appendix). The full experiment has 8--20M samples. \sys maintains its high performance, even with as little as $n_b=500$ samples, but drops below this point (statistically significant in over one third of experiments). This is not surprising, as the context traces do not change drastically at short intervals, and even 500 randomly sampled points from the trace should be enough to have a representation over all of the trace. However, with more complicated and high-dimensional contexts, a higher buffer size would likely be necessary. 

\section{Discussion and Limitations}
\label{sec:limits}

\paragraph{Network Capacity.}
In general, online learning methods with bounded parameter counts will reach the function approximator's (neural network's) maximum representational capacity. \sys is not immune from this, as we do not add parameters with more context traces. However, neither are prescient agents. To isolate the effect of this capacity and \gls{cf}, we compare against prescient agents, rather than single agents trained on individual tasks or context traces~\citep{he2020task}. This ensures a fair evaluation that does not penalize online learning for reaching the capacity ceiling. If the maximum capacity has been reached, it may be beneficial to remove significantly old samples from $B_a$ to allow \sys to forget such contexts, thereby favoring flexibility instead of stability.

\paragraph{Exploration.}
\sys focuses on mitigating catastrophic forgetting in non-stationary \gls{rl}. An orthogonal challenge in this setting is efficient exploration, i.e., to explore when the context distribution has changed but only once per new context. Our experiments used automatic entropy tuning for exploration~\citep{haarnoja2018soft}; while empirically effective, this was not designed for non-stationary problems. \sys may benefit from a better exploration methodology such as curiosity-driven exploration~\citep{pathak2017curiosity}. 

\paragraph{Efficient Buffer Management.}
We used Reservoir Sampling~\citep{vitter1985random}, which maintains a uniformly random buffer of all observed state-context samples so far. Future work could explore strategies that selectively store or drop samples based on their context, e.g., to maximize sample diversity.
\section{Conclusion}
\label{sec:conclusion}

We proposed and evaluated \sys, a simple approach for online learning in non-stationary context-driven environments. \sys requires two conditions: (1) the non-stationarity must be induced by an exogenous observed context process; and (2) a similarity metric is required that can inform us if two contexts come from noticeably different distributions (\gls{ood} detection). This is less restrictive than prior approaches that require either explicit or inferred task labels. Our experiments showed that \sys outperforms baselines on several environments with real and synthetic context processes. 

\section*{Reproducibility Statement}

For \Cref{theorem:tabular_episodic}, we include the proof and assumptions in \S\ref{sec:theory}. We include detailed accounts of environments, context traces, baselines, hyperparameters, software and hardware in \S\ref{sec:eval} in the main text and \S\ref{sec:pend_env} and \S\ref{sec:lbalance_env} in the Appendix. We include implementation details and pseudo algorithms in \S\ref{sec:method} and \S\ref{sec:implementation} in the Appendix. A link to the source code is also provided in \S\ref{sec:intro}.
\section*{Acknowledgements}

We thank our reviewers for insightful comments. This work was supported by NSF grants 1751009 and an award from the CSAIL-MSR Trustworthy AI collaboration.
\newpage

\bibliography{paper}
\bibliographystyle{iclr2025_conference}

\clearpage
\appendix
\begin{appendices}
\appendix
\section{Related Work: Continued}
\label{sec:related_works_cont}

\paragraph{Meta-learning.}
Here, in the general case, the environment can switch between a set of possible \glspl{mdp} without an explicit signal. At the `meta-train' phase, the policy is allowed to train on a part or all of the \glspl{mdp}. At the `meta-test' phase, the policy must make decisions in a continual \gls{rl} setup where the \gls{mdp} may abruptly change, and it has to adapt to the current \gls{mdp} with few-shot adaptation~\citep{alshedivat2018continuous,nagabandi2019deep,nagabandi2019learning} or context inference~\citep{rakelly2019efficient}. In our problem setup, we assume no access to the environment beforehand. 

\paragraph{Multi-task \gls{rl}.}
In this problem, there are several (e.g., 10 or 50) different tasks with different \glspl{mdp}. The goal of this line of work is to learn a shared policy for all tasks that approaches the performance of learning separate policies per task~\citep{yang2020multitask,sodhani2021multitask}. These tasks may come with contextual information about the task that can be used in the policy~\citep{sodhani2021multitask}. This problem is not continual \gls{rl}, does not experience \gls{cf}, and the learner is allowed to explore all tasks at the same time. Often the goal is postitive transfer learning, i.e., learning faster on all tasks in parallel than learning tasks separately~\citep{teh2017distralrobustmultitaskreinforcement}.

Another type of work focuses on speeding up the learning process for a set of new tasks by pre-training on a set of old tasks~\citep{xue2024stateregularizedpolicyoptimization}. This line of work bears similarities to multi-task \gls{rl} and meta-learning.

Assuming there are no explicit signals for environment contexts, \citet{wei2021nonstationary} provide a regret-optimal black-box \gls{rl} algorithm. 

\paragraph{Interfernce in stationary \gls{rl}.}
Investigating interference in vanilla RL (non-stationarity in sample distribution) is an adjacent and interesting line of work~\citep{bengio2020interferencegeneralizationtemporaldifference, pan2021fuzzytilingactivationssimple, liu2023measuringmitigatinginterferencereinforcement, liu2018utilitysparserepresentationscontrol}. The type of non-stationarity discussed in these works is different than what we study. Here, non-stationarity refers to the moving target of bootstrap losses, such as Q-learning, due to shifting policies that change future data distributions. Non-stationarity in our problem setup means the MDP itself is shifting, irrespective of the changes the policy makes. The second type of non-stationarity cannot be resolved even with "perfect" RL algorithms that deal with interference in stationary MDPs.

However, there may be ideas that are transferable. For example, this line of work suggests that techniques such as \gls{gae}~\citep{schulman2018highdimensional} and target networks reduce interference~\citep{bengio2020interferencegeneralizationtemporaldifference, liu2023measuringmitigatinginterferencereinforcement}. Another interesting avenue is experimentation with sparse learning~\citep{pan2021fuzzytilingactivationssimple, liu2018utilitysparserepresentationscontrol}.

\paragraph{Buffer limitation interference.}
A line of work, related to interference, deals with a type of non-stationarity induced by having small buffers (e.g., 32 samples, compared to the typical thousands to millions) in off-policy algorithms. These works aim to mimic an off-policy agent with unbounded buffers, and do not focus on context-driven non-stationarity. They learn policies that perform closely to unbounded agents via techniques such as following old target Q networks~\citep{lan2023memoryefficientreinforcementlearningvaluebased} or utilizing sparse-gradient activation functions~\citep{lan2023elephantneuralnetworksborn}. Note that we do compare to DDQN and SAC with unbounded buffers.
\section{Proof of Optimality in Tabular Context-driven RL}
\label{sec:theory}

Below, we show how the non-stationary environment in \S\ref{sec:background} can be learned with vanilla \gls{rl} algorithms, if the state, action and context spaces are finite, context switches occur at episode boundaries. First, we prove Lemma \S\ref{lemma:learn_rates}, and then we prove the main theorem \Cref{theorem:tabular_episodic}.

\begin{lemma}
\label{lemma:learn_rates}
    Given a monotonically decreasing sequence $\{\alpha_i\}_{i=1}^{\infty}$ that satisfies the following conditions:
    \begin{equation}
        \sum_{i=1}^{\infty} \alpha_i = \infty\qquad \sum_{i=1}^{\infty} \alpha_i^2 < \infty\\
    \end{equation}
    
    Consider any subsequence $\{\beta_j\}_{j=1}^{\infty}$, where for any $1 \le j$, there exists $N \times (j-1) < i \le N \times j$ such that $\beta_j = \alpha_i$.
    
    Prove that:
    \begin{equation}
        \sum_{j=1}^{\infty} \beta_j = \infty\qquad \sum_{j=1}^{\infty} \beta_j^2 < \infty\\
    \end{equation}
\end{lemma}

\begin{proof}
    First, note that since $\{\alpha_i\}_{i=1}^{\infty}$ is monotonically decreasing, we have:
    \begin{equation}
        \alpha_{N \times i} \le \beta_i \le \alpha_{N\times (i-1)+1}\\
    \end{equation}

    For the first equality, we have for all $k > 0$:
    \begin{equation}
        \begin{aligned}
            \sum_{i=1}^{\infty} \beta_i &\ge \sum_{i=1}^{\infty} \alpha_{N \times i}\\
            &\ge \sum_{i=1}^{\infty} \alpha_{N \times i + k}\\
        \end{aligned}
    \end{equation}
    
    Thus, we have:
    \begin{equation}
        \begin{aligned}
            N \times \sum_{i=1}^{\infty} \beta_i &= \sum_{k=1}^{N} \sum_{i=1}^{\infty} \beta_i\\
            &\ge \sum_{k=1}^{N} \sum_{i=1}^{\infty} \alpha_{N \times i+k}\\
            &= \sum_{i=N+1}^{\infty} \alpha_{i}\\
        \end{aligned}
    \end{equation}
    
    Therefore:
    \begin{equation}
        \begin{aligned}
            \sum_{i=1}^{\infty} \beta_i &\ge \frac{1}{N} \sum_{i=1}^{\infty} \alpha_{i} - \frac{1}{N} \sum_{i=1}^{N} \alpha_{i} = \infty\\
        \end{aligned}
    \end{equation}
    
    The second bound is trivially proven:
    \begin{equation}
        \begin{aligned}
            \sum_{i=1}^{\infty} \beta_i^2 &\le \sum_{i=1}^{\infty} \alpha_{N\times (i-1)+1}^2\\
            &\le \sum_{i=1}^{\infty} \alpha_{i}^2\\
            &< \infty\\
        \end{aligned}
    \end{equation}
\end{proof}

\begin{theorem}
\label{theorem:tabular_episodic}
    Consider a context-driven \gls{mdp} as defined in \S\ref{sec:background}. Under the following set of assumptions, prove that the Q-learning algorithm~\citep{watkins1992q} converges to the optimal policy:
    \begin{enumerate}
        \item Rewards are bounded bounded rewards $|r_t|\ge R$.
        \item We have a a monotonically decreasing sequence of learning rates $\{\alpha_i\}_{i=1}^{\infty}$ where $0 \ge \alpha_i < 1$, and
        \begin{equation}
            \sum_{i=1}^{\infty} \alpha_i = \infty\qquad \sum_{i=1}^{\infty} \alpha_i^2 < \infty\\
        \end{equation}
        \item The state, action and context spaces are finite $|\mathcal{S}|, |\mathcal{A}|, |\mathcal{Z}|<\infty$. 
        \item There exists $N$, such that for any context $z\in \mathcal{Z}$, $z$ occurs at least once in any consecutive subsequence of the context trace of size $N$.
        \item The context only changes at episode terminations.
    \end{enumerate}
\end{theorem}

\begin{proof}
    As contexts only switch at episode boundaries, the Q-learning updates for each context $z\in \mathcal{Z}$ are isolated from Q-values of other contexts. Thus, for any context $z\in \mathcal{Z}$, we have a separate Q-learning session. If $I_{z}$ denotes the set of episode indices where the context trace is equal to $z$, If we prove
    \begin{equation}
        \sum_{i \in I_z} \alpha_i = \infty \qquad \sum_{i \in I_z} \alpha_i^2 < \infty\\
    \end{equation}
    
    we can use the original Q-learning proof of convergence~\citep{watkins1992q}.
    
    To prove this, note that from assumption 5 we can surmise that that for any $1 \le j$, there exists $N \times (j-1) < i \le N \times j$ such that $i \in I_z$. Therefore based on Lemma \S\ref{lemma:learn_rates} the subsequence of learning rates satisfy the Watkins Q-learning condition. 
\end{proof}
\section{\sys Implementation}
\label{sec:implementation}

\begin{figure}[!t]
    \centering
    \includegraphics[width=0.8\linewidth]{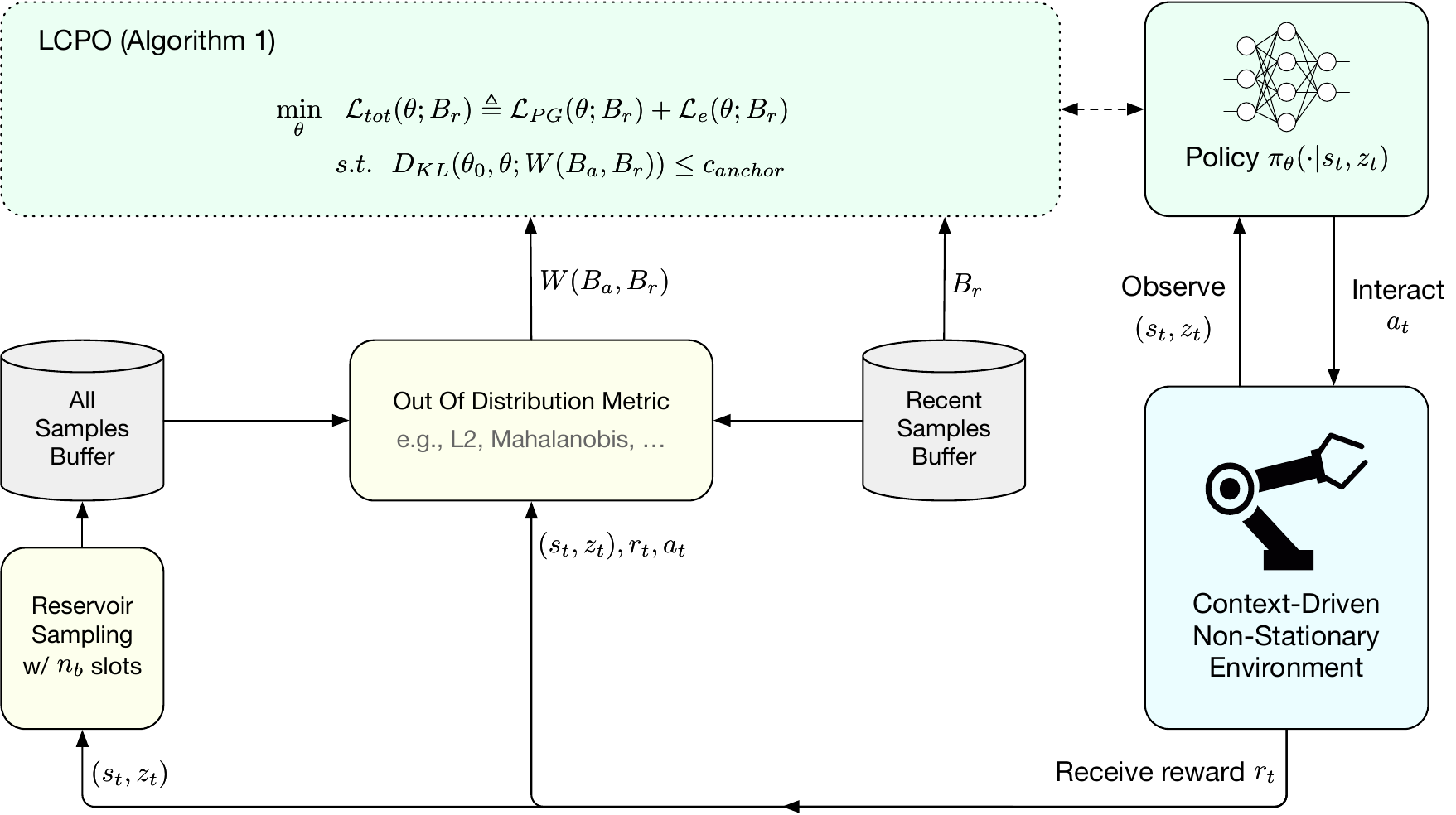}
    \caption{Architecture of \sys.}
    \label{fig:lcpo_block_diagram}
\end{figure}

Here, we will discuss the implementation details of \sys. \Cref{fig:lcpo_block_diagram} depicts the overall architecture of \sys.

\subsection{Reservoir Sampling}
\label{sec:lcpo_reservoir}

As discussed in \S\ref{sec:method}, \sys needs to maintain a buffer $B_a$ of all samples observed so far. However, this buffer will grow with time and incur extensive memory costs. To limit the buffer size while maintaining a distribution of all samples observed so far, we utilize Reservoir Sampling~\citep{vitter1985random}.

The pseudo-code for the implementation can be found in \Cref{algo:reserv}. Reservoir sampling operates by maintaining a bounded list of samples $B_a$. While the number of samples in the list $B_a$ has not reached max capacity $n_b$, all samples are admitted. Once the buffer is full, a random index is sampled in the range of 0 to $n_s-1$, where $n_s$ is the number of samples observed so far. If the index is smaller than $n_b$, the element at index $i$ is replaced with the new sample. If it is larger, the sample is thrown away. This strategy bounds the size of $B_a$, but maintains a uniform distribution of samples from the true list of all samples observed so far.

\begin{algorithm}[ht]
\caption{Reservoir Sampling}
\label{algo:reserv}
\begin{algorithmic}[1]
\State \textbf{Input:} max capacity $n_b$
\State $B_a \gets \{\}$ Initialize empty buffer
\State $n_s \gets 0$ Initialize sample count
\For{each new sample $x$}
    \State $n_s \gets n_s + 1$ Increment the sample count
    \If {$n_s > n_b$}
        \State $i \sim Unif(0, n_s-1)$ Sample a random index smaller than $n_s$
        \If {$i < n_b$}
            \State Replace element at index $i$ in $B_a$ with $x$
        \Else
            \State Throw new sample $x$ away
        \EndIf
    \Else
        \State $B_a \gets B_a + \{x\}$ Append $x$ to $B_a$
    \EndIf
\EndFor
\end{algorithmic}
\end{algorithm}

\subsection{OOD Function}
\label{sec:lcpo_ood_dist}

\sys requires a definition for \gls{ood} samples, i.e., samples that come from contexts far away from recent samples $B_r$. Intuitively, the optimal policy conditioned on this \gls{ood} context should be significantly different from the policy optimal conditioned on the current context.

Such metrics can be based on domain insight, where an expert who is familiar with how the context value changes the \gls{mdp} would define a function to detect \gls{ood} samples. Alternatively, it may be generic \gls{ood} criteria commonly used in the literature, such as L2 thresholding, Mahalanobis distance, etc. In this case, the \gls{ood} function needs to be tuned to be meaningful, which is a common challenge in \gls{ood} detection work. An interesting line of future work is to utilize \gls{mdp} transitions in the warm-up period for tuning the threshold online.

Concretely, \sys requires an \gls{ood} function $\omega(z, B_r)$ that denotes whether $z$ is \gls{ood} with respect to $B_r$ or not. In the L2 thresholding used for the gym environments subject to the wind context in \S\ref{sec:eval}, we calculate an average over $B_r$, i.e., $\mu_w=\mathbb{E}_{w \sim B_r}[w]$, and define $\omega(w, B_r) := \lVert w - \mu_w \rVert_2 > \sigma$ for some threshold $\sigma$. For the Mahalanobis distance~\citep{mahalanobis2018generalized} measure used for the straggler mitigation experiments in S\ref{sec:eval}, we calculate an average and standard deviation over $B_r$, i.e., $\mu_{w}=\mathbb{E}_{w \sim B_r}[w]$ and $\Sigma_w=\mathbb{E}_{w \sim B_r}[(w-\mu_w)^2]$, and define $\omega(w, B_r) := (w - \mu_w)^T\Sigma_w(w - \mu_w) > \sigma^2$ for some threshold $\sigma$.

\subsection{OOD Sampling}
\label{sec:lcpo_ood_sample}

Finally when want to sample a batch of size $b$ from $W(B_a, B_r)$, forming the full set $W(B_a, B_r)$ and then sampling randomly from it has a computational cost that scales with $|B_a|$. To avoid this cost, we instead sample $s$ experiences from $B_a$ and keep the ones that are \gls{ood}. If this results in at least $b$ \gls{ood} experiences, we return the samples. If not, we conclude that there aren't enough OOD samples in $B_a$ with respect to $B_r$. A pseudo-code is provided in \Cref{algo:w_sample}.

Analytically we can model the sampling with a binomial distribution, where $p=\mathbb{E}_{z \sim B_a}[\omega(z, B_r)]$ is the fraction of samples in $B_a$ that are \gls{ood} with respect to $B_r$. We will successfully get a batch of \gls{ood} samples with the probability $1-F(b;s,p)$ where $F(\cdot;\cdot,\cdot)$ is the binomial cumulative distribution function. In all experiments in \S\ref{sec:eval} we have set $s=5b$. With $b=200$, the success probability is 5\% when $p=0.18$ and 94\% when $p=0.22$. In other words, if at least 22\% of the samples in $B_a$ are \gls{ood}, we are highly likely to be able to collect a batch of size $b$ of \gls{ood} samples, and unlikely if the rate is 18\% or below.

\begin{algorithm}[h]
\caption{OOD Sampler}
\label{algo:w_sample}
\begin{algorithmic}[1]
\State \textbf{Input:} All samples buffer $B_a$, Recent samples buffer $B_r$, OOD function $\omega(z, B_r)$, $s$ max sample count, $b$ batch size
\State Initialize empty list $B_{out} \gets \{\}$
\State $i \gets 0$
\While{$\left|B_{out}\right| < b$ and $i < s$}
    \State $z \sim B_a$ Sample from $B_a$
    \If {$\omega(z, B_r)$ Sample is OOD}
        \State $B_{out} \gets B_{out}+\{z\}$ Add sample to list
    \EndIf
    \State $i \gets i + 1$ Increment $i$
\EndWhile
\If {$\left|B_{out}\right| = b$}
    \State return $B_{out}$
\Else
    \State return $\{\}$
\EndIf
\end{algorithmic}
\end{algorithm}

\section{Baselines}
\label{sec:baselines}

\subsection{Online EWC}
\label{subsec:base_ewc}

\gls{ewc}~\citep{kirkpatrick2017overcoming} regularizes online learning with a Bayesian approach assuming task indices, and online \gls{ewc}~\citep{chaudhry2018riemannian, schwarz2018progress} generalizes it to task boundaries. To adapt online \gls{ewc} to our problem, we update the importance vector and average weights using a weighted moving average in every time step. The underlying learning approach is \gls{sac}~\citep{haarnoja2018soft}.

\gls{ewc} applies a regularization loss 
\begin{equation}
    \mathcal{L}_{ewc}=\alpha\sum_{k=1}^{N}\lvert\lvert \theta_t - \theta_{k}^* \rvert\rvert^2_{F_k}
\end{equation}
to the training loss, where $N$ are the number of tasks, $\theta_{k}^*$ is the converged parameter set for task $k$ and $F_k$ is the diagonal of the Fisher Information Matrix (FIM) of task $k$ on the converged model for task $k$. Online \gls{ewc} applies an approximation of this regularization loss
\begin{equation}
    \mathcal{L}_{ewc}=\alpha\lvert\lvert \theta_t - \theta_{t-1}^* \rvert\rvert^2_{F_{t-1}^*}
\end{equation}
using a running average $F_t^*$ of the diagonal of the Fisher Information Matrix (FIM), and a running average of the parameters $\theta_t^*$. The running average $F_t^*$ is updated with a weighted average $F_t^*=(1-\beta)F_{t-1}^*+\beta F_t$, where $F_t$ is the diagonal of the FIM respective to the recent parameters and samples.\footnote{We deviate from the original notations $\{\lambda, \gamma\}$~\citep{rusu2016progressive}, since they could be confused with the MDP discount factor $\gamma$ and GAE discount factor $\lambda$~\citep{schulman2018highdimensional}.} Similarly, the running average $\theta_t^*$ uses the same parameter $\beta$.

\begin{figure}[ht]
    \centering
    \includegraphics[width=\linewidth]{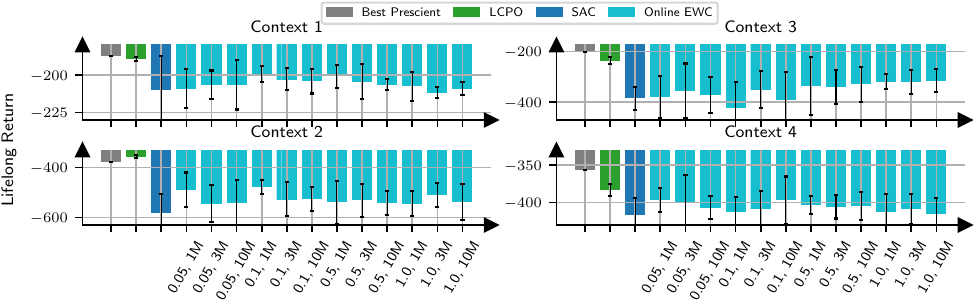}
    \caption{
    Pendulum-v1 lifelong returns and 95\% confidence bounds of Online EWC with 12 hyperparameter trials. Hyperparameters are labeled as $\{\alpha, \beta^{-1}\}$, where $\alpha$ is the regularization strength and $\beta$ is the averaging weight. The optimal online EWC hyperparameter is sensitive to the context, but \sys is better even if online \gls{ewc} is tuned per context.}
    \label{fig:windy_pend_ewc_returns}
\end{figure}

Online \gls{ewc} may constrain the policy output to remain constant on samples in the last $\approx\beta^{-1}$ epochs, but it has to strike a balance between how fast the importance metrics are updated with the newest FIM (larger $\beta$) and how long the policy has to remember its training (smaller $\beta$). This balance will ultimately depend on the context trace and how frequently they evolve. We tuned $\alpha$ and $\beta$ on Pendulum-v1 for all contexts, trying $\alpha \in \{0.05, 0.1, 0.5, 1.0, 10, 100, 1000\}$ and $\beta^{-1} \in \{1M, 3M, 10M\}$ (M denotes 1 million). The returns are visualized in \Cref{fig:windy_pend_ewc_returns} with full details in \Cref{table:ewc_return}. There is no universal $\beta$ that works well across all contexts and online \gls{ewc} would not perform better than \sys even if tuned to each context trace individually. We ultimately chose $\beta^{-1}=3M$ samples to strike a balance across all contexts, but it struggled to even surpass \gls{sac} on other environments. 

\begin{table*}[ht]
\caption{Average and 95th percentile confidence ranges for lifelong returns for online \gls{ewc} variants in the Pendulum-v1 environment with external wind processes.}
\label{table:ewc_return}
\scriptsize
\centering
\begin{tabular}{l c c c @{}c c c c @{}c c c c @{}c c c c}
\toprule
 & \multicolumn{14}{c}{\textbf{Online EWC}}\\
\cmidrule{2-16}
 & \multicolumn{3}{c}{$\alpha=0.05$} & & \multicolumn{3}{c}{$\alpha=0.1$} & & \multicolumn{3}{c}{$\alpha=0.5$} & & \multicolumn{3}{c}{$\alpha=1.0$}\\
\cmidrule{2-4} \cmidrule{6-8} \cmidrule{10-12} \cmidrule{14-16}
$\beta^{-1}$: & $1M$ & $3M$ & $10M$ & & $1M$ & $3M$ & $10M$ & & $1M$ & $3M$ & $10M$ & & $1M$ & $3M$ & $10M$ \\

\midrule
Context & -209 & -207 & -206 & & -199 & -203 & -204 & & -201 & -204 & -207 & & -208 & -212 & -209 \\
Trace 1 & $\pm$13.0 & $\pm$9.59 & $\pm$16.6 & & $\pm$5.47 & $\pm$7.28 & $\pm$8.06 & & $\pm$7.57 & $\pm$11.5 & $\pm$3.67 & & $\pm$9.69 & $\pm$3.87 & $\pm$4.49 \\
\midrule
Context & -489 & -545 & -544 & & -477 & -527 & -525 & & -540 & -531 & -541 & & -545 & -510 & -539 \\
Trace 2 & $\pm$68.6 & $\pm$74.1 & $\pm$95.3 & & $\pm$26.9 & $\pm$67.9 & $\pm$47.6 & & $\pm$84.6 & $\pm$65.9 & $\pm$48.0 & & $\pm$49.0 & $\pm$48.5 & $\pm$71.2 \\
\midrule
Context & -380 & -356 & -372 & & -422 & -349 & -393 & & -335 & -340 & -330 & & -320 & -320 & -316 \\
Trace 3 & $\pm$82.4 & $\pm$108 & $\pm$70.0 & & $\pm$102 & $\pm$72.4 & $\pm$111 & & $\pm$114 & $\pm$66.1 & $\pm$68.5 & & $\pm$29.1 & $\pm$46.1 & $\pm$45.7 \\
\midrule
Context & -396 & -400 & -407 & & -413 & -408 & -397 & & -404 & -406 & -404 & & -412 & -409 & -416 \\
Trace 4 & $\pm$16.1 & $\pm$36.7 & $\pm$15.6 & & $\pm$21.1 & $\pm$23.0 & $\pm$31.9 & & $\pm$11.7 & $\pm$15.4 & $\pm$18.5 & & $\pm$23.5 & $\pm$20.1 & $\pm$21.7 \\

\bottomrule
\end{tabular}
\end{table*}

\subsection{Sliding OGD}
\label{subsec:base_ogd}

\gls{ogd}~\citep{farajtabar2019orthogonal} projects gradients for new tasks to vector spaces that are orthogonal to previous tasks' loss functions. \gls{ogd} needs task labels, and \citet{woo2022structure} circumvent this by making projecting.

\gls{ogd}~\citep{farajtabar2019orthogonal} circumvents \gls{cf} by applying parameter updates that are orthogonal to the losses of past tasks. To do this, after a task has finished training, \gls{ogd} calculates the gradient vector w.r.t.  to samples for that task and stores those gradients. When the next task training begins, gradient updates are projected to orthogonal spaces w.r.t. the saved vectors from before. This procedure requires task labels and convergence. Sliding \gls{ogd}~\citep{woo2022structure} avoids needing task labels by using the gradient updates in the past $N$ episodes for projection. In other words, Sliding \gls{ogd} implicitly assumes that each of the past $N$ episodes were ``tasks'' that have already finished training. This assumption is incorrect in our problem setup, as the context trace may change slowly. As a result, the gradient vectors of past episodes belong to the same ``task'' as the current episodes, and this projection fully hinders training.

\subsection{Benna Fusi DQN}
\label{subsec:base_bfdqn}

This approach applies a biologically plausible model for synapses on neural network weights in a deep Q network~\citep{kaplanis2018continual}. Conceptually, the weights are regularized with their past values in multiple different time scales. \citet{kaplanis2018continual} note that while the Benna-Fusi DQN architecture was successful in simple environments, it failed with more complex and challenging ones. In our experience, this architecture did not provide any benefits compared to vanilla \gls{ddqn}.

\subsection{MBCD}
\label{subsec:base_mbcd}

This work handles piecewise stationary environments by inferring change-points and task labels~\citep{alegre2021minimum}. It trains models to predict environment state transitions, and launches new policies when the current model is inaccurate in predicting the state trajectory based on the CUSUM algorithm~\citep{page1954continuous}. The underlying learning approach is \gls{sac}~\citep{haarnoja2018soft}.

\gls{mbcd}'s sensitivity for detecting environment changes is a tunable hyperparameter; we tuned it by trying 6 values in a logarithmic space spanning $10^1$ to $10^6$ on the evaluation context traces with Pendulum-v1, and chose the best performing hyperparameter on the test set. \gls{mbcd} still endlessly spawned new policies for other environments, and therefore we limited the maximum number of models to 7. Despite this, \gls{mbcd} fails to perform well over the diverse set of contexts. \gls{mbcd} struggles to tease out meaningful task boundaries. In some experiments, it launches anywhere between 3 to 7 policies just by changing the random seed. This observation is in line with \gls{mbcd}'s sensitivity in \S\ref{sec:motivation}.

\subsection{MBPO}
\label{subsec:base_mbpo}

MBPO~\citep{janner2021trust} is a model-based approach that trains an ensemble of experts to learn the environment model, and generates samples for an \gls{sac}~\citep{haarnoja2018soft} algorithm. If the model is accurate, it can fully replay prior contexts, thereby avoiding catastrophic forgetting. 

\gls{mbpo} performed poorly. If the fault is the accuracy of the learned environment models, it could be improved with approaches such online meta-learning~\citep{finn2019online} or goal-oriented model-based learning~\citep{pong2020temporal}.. We investigated the learned models for the Pendulum-v1 experiments manually. We found these models to be very accurate, since the environment dynamics are simple.

To concretely verify that the accuracy of \gls{mbpo} learned models is not the reason it underperforms, we instantiated an \gls{mbpo} agent with access to the ground truth environment dynamics and context traces (but not future context traces), which we call {\em Ideal \gls{mbpo}}. We compare the performance of Ideal \gls{mbpo} vs. standard \gls{mbpo} in \Cref{table:imbpo_return}. The performance of these two agents is widely similar. This confirms the fact that the learning algorithm itself is the problem, and not the learned models.

\begin{table*}[h]
\caption{Average and 95th percentile confidence ranges for lifelong returns for different algorithms and conditions in the Pendulum-v1 environment with external wind processes. An MBPO agent with access to the ground truth model performs similarly to the MBPO model. Schemes with superiority beyond 95\% confidence are highlighted in bold.}
\label{table:imbpo_return}
\scriptsize
\centering
\begin{tabular}{l c c c c c}
\toprule
 & \multicolumn{4}{c}{\textbf{Online Learning}} & \\
\cmidrule{2-5}
& LCPO & MBPO & Ideal MBPO & A2C & \textbf{Best Prescient}\\

\midrule
\multirow{2}{*}{Context Trace 1} & \textbf{-190} & -325 & -320 & -201 & \multirow{2}{*}{-187 (SAC)} \\
 & \textbf{$\pm$0.45} & $\pm$47.2 & $\pm$44.2 & $\pm$3.92 & \\
\midrule
\multirow{2}{*}{Context Trace 2} & \textbf{-355} & -843 & -870 & -377 &  \multirow{2}{*}{-376 (DDQN)} \\
 & \textbf{$\pm$2.57} & $\pm$24.2 & $\pm$56.4 & $\pm$8.05 &   \\
\midrule
\multirow{2}{*}{Context Trace 3} & \textbf{-240} & -603 & -637 & -307 &  \multirow{2}{*}{-203 (SAC)} \\
 & \textbf{$\pm$4.41} & $\pm$213 & $\pm$90.0 & $\pm$32.6 &   \\
\midrule
\multirow{2}{*}{Context Trace 4} & \textbf{-378} & -553 & -555 & -399 & \multirow{2}{*}{-357 (SAC)} \\
 & \textbf{$\pm$3.02} & $\pm$90.9 & $\pm$84.1 & $\pm$7.76 &   \\

\bottomrule
\end{tabular}
\end{table*}

The reason is the way that \gls{mbpo} samples experiences for training. At every iteration, MBPO samples a batch of actual interactions from its experience buffer and generates hypothetical interactions from them. These hypothetical interactions amass in a second buffer, which is used to train an \gls{sac} agent. During the course of training, the second buffer accumulates more interactions generated from samples from the start of the experiment compared to recent samples. This is not an issue when the problem is stationary, but in non-stationary \gls{rl} subject to an context process, this leads to over-emphasis of the context processes encountered earlier in the experiment. As such, \gls{mbpo} fails to even surpass \gls{sac}. Prior work has observed the sensitivity of off-policy approaches to such sampling strategies~\citep{isele2018selective, hamadanian2022demistify}.

\subsection{CLEAR}
\label{subsec:base_clear}

This approach aims to mitigate \gls{cf} with off-policy learning and maintain quick adaptation with on-policy learning~\citep{rolnick2019experience}. They use IMPALA and V-trace~\citep{espeholt2018impala} on recent batches for on-policy and stale batches for off-policy \gls{rl}.

CLEAR~\citep{rolnick2019experience} aims to mix the quick adaptation of on-policy \gls{rl} and the \gls{cf} resilience of off-policy \gls{rl}, but in practice this fusion makes it very hyperparameter sensitive. The V-trace algorithm was originally intended to correct for lagging policies in a distributed \gls{rl} architecture~\citep{espeholt2018impala}. V-trace uses clipped importance sampling to correct for the drift between the logging and training policies, which reduces variance but biases the loss. With small lags between workers, the bias is small. If V-trace is used in a scenario where the logging and training policy are very different, such as in CLEAR~\citep{rolnick2019experience}, the bias becomes significant enough to hinder training. CLEAR attempts to circumvent this by inducing a regularization loss on the actor and critic. Yet, this regularization will count against improving the \gls{rl} policy for better returns, and will be brittle. The correct balance between policy improvement and this regularization will ultimately depend on the environment and context trace. While we tuned CLEAR extensively for the Pendulum-v1 environment, as we did for all baselines, it fails catastrophically in other environments.

\subsection{PT-DQN}
\label{subsec:base_ptdqn}

PT-DQN~\citep{anand2023predictioncontrolcontinualreinforcement} learns two separate networks with two different goals; (1) a permanent network that is updated infrequently and slowly, and aims to learn a generalized estimate of Q-values in various tasks, and (2) a transient network that learns \emph{and forgets} aggressively, and aims to quickly learn the optimal policy for the current task. \citet{anand2023predictioncontrolcontinualreinforcement} prove PT-DQN asymptotically converges to the optimum predictors in piecewise stationary prediction tasks with tabular input spaces.

Note that PT-DQN avoids catastrophic forgetting indirectly by slowly updating a permanent Q-network. The hope is that it strikes the right balance in learning this network slowly enough such that earlier contexts are not forgotten, but updates it frequently enough such that new knowledge is not lost. This trade-off is brittle; how fast the transient Q-network should forget and relearn, and how slowly the permanent Q-network should be updated highly depends on (1) how quickly the context process changes, and (2) by how much these changes affect the transition dynamics of the MDP. As was also observed with CLEAR (\S\ref{subsec:base_clear}), balances of this nature are brittle, due to the indirect nature of how these techniques address catastrophic forgetting.

Thus, PT-DQN is understandly hyperparameter sensitive. Indeed, to tune PT-DQN on Pendulum-v1 environment, (as done for all baselines), we carried out three rounds of grid-search on five hyperparameters, totalling 440 different combinations. Despite PT-DQN being competitive with DDQN on Pendulum-v1, the performance is unpredictable in other environments.

\subsection{Off-policy RL}
\label{subsec:base_off_rl}

Off-policy \gls{rl} is potentially capable of overcoming \gls{cf} due to replay buffers, at the cost of unstable training. We consider \gls{ddqn}~\citep{doubledqn} and \gls{sac} (with automatic entropy regularization, similar to \sys)~\citep{haarnoja2018soft}. 

\subsection{On-policy RL}
\label{subsec:base_on_rl}

On-policy \gls{rl} is susceptible to \gls{cf}, but more stable in online learning compared to off-policy \gls{rl} algorithms, and the fast adaptation of these algorithms can also help them `track' the optimal policy as the environment changes~\citep{sutton2007tracking}. We compare with \gls{a2c}~\citep{mnih2016asynchronous} and \gls{trpo} (single-path)~\citep{schulman2015trust}, with \gls{gae}~\citep{schulman2018highdimensional} applied. Note that \gls{trpo} vine is not possible in online learning, as it requires rolling back the environment world. 

\subsection{Prescient RL}
\label{subsec:prescient}

To establish an upper-bound on the best performance that an online learner can achieve, we train \emph{prescient policies}, as discussed in \S\ref{sec:background}. We allow prescient policies to have unlimited access to the contexts and environment dynamics, i.e., they are able to replay any particular environment and context as many times as necessary. Since prescient policies can interact with multiple contexts in parallel during training, they do not suffer from \gls{cf}. In contrast, all other baselines (and \sys) are only allowed to experience contexts sequentially as they occur over time and must adapt the policy on the go. We report results for the best of four prescient policies with the following model-free algorithms: \gls{a2c}~\citep{mnih2016asynchronous}, \gls{trpo} (single-path)~\citep{schulman2015trust}, \gls{ddqn}~\citep{doubledqn} and \gls{sac}~\citep{haarnoja2018soft}.
\section{\sys variants}
\label{sec:lcpo_variants}
In \S\ref{sec:method} we discussed our main approach to solving the constrained optimization problem below:

\begin{equation}
    \label{eqn:lcpo_s_pure}
    \centering
    \begin{aligned}
        \min_{\theta}~~ \mathcal{L}_{PG}(\theta; B_r)& \\
        s.t.~~ D_{KL}(\theta_0, \theta; W(B_a, B_r&)) \le c_{anchor} \\
    \end{aligned}
\end{equation}

The goal is to optimize the policy gradient loss, i.e., maximize returns on current input, while minimizing policy change on past observed inputs with the \emph{anchoring constraint}. We outlined a solution to this problem, with a pseudo-code in \Cref{algo:train}, that is essentially a second order constrained optimization plus a line search phase:

\begin{equation}
    \begin{aligned}
        \min_{\theta}~~ (\theta-\theta_0)^T &v_{PG} \\
        s.t.~~ (\theta-\theta_0)^T A(\theta-&\theta_0) \le c_{anchor} \\
    \end{aligned}
\end{equation}

where $A_{ij}=\frac{\partial}{\partial \theta_i}{\frac{\partial}{\partial \theta_j}} D_{KL}(\theta_0, \theta; W(B_a, B_r))$, and $v_{PG}=\nabla_{\theta}\mathcal{L}_{PG}(\theta; \cdot)\vert_{\theta_0}$. However, there is another way to solve this problem, that we discuss as follows.

\subsection{\sys-P}
\label{sec:lcpo_p}

An alternative way to uphold the anchoring constraint is to directly add it as term in the loss function. Let us define:

\begin{equation}
    \begin{aligned}
        \mathcal{L}_{anchor}(\theta, \theta_0; B_r, B_a) =& \\
        \mathbb{E}_{s, z \sim W(B_a, B_r)}[CEL&oss(\pi_{\theta_0}(s, z), \pi_{\theta}(s, z))] \\
    \end{aligned}
\end{equation}

Where we use the Cross Entropy loss to incentivize policy anchoring. Then, we optimize the following total loss:

\begin{equation}
    \begin{aligned}
        \min_{\theta}~~ &\mathcal{L}_{PG}(\theta; \cdot) + \kappa.\mathcal{L}_{anchor}(\theta, \theta_0; B_r, B_a)
    \end{aligned}
\end{equation}

This approach is even less compute intensive than \sys, but is not possible in vanilla policy gradient. This is because the gradient direction from $\mathcal{L}_{anchor}$ is zero when $\theta=\theta_0$ and will not affect the optimization. Therefore, we have to repeat the gradient update several times before this term has an effect. The optimization setup in \gls{ppo}~\citep{schulman2017proximal} allows for several gradient steps with one batch of data, and therefore we apply the above loss in the \gls{ppo} optimization step. We dub this approach \sys-P (P stands for proximal).

\begin{table*}[h]
\caption{Average and 95th percentile confidence ranges for lifelong returns for \sys and \sys-P and conditions in the Pendulum-v1 environment with external wind processes. Schemes with superiority beyond 95\% confidence are highlighted in bold.}
\label{table:windypend_lcpo_vars}
\scriptsize
\centering
\begin{tabular}{l c c c c}
\toprule
& LCPO & LCPO-P & Best Baseline & \textbf{Best Prescient} \\

\midrule
\multirow{2}{*}{Context Trace 1} & \textbf{-190} & -211 & -201 (A2C) & \multirow{2}{*}{-187 (SAC)} \\
 & \textbf{$\pm$0.45} & $\pm$3.92 & $\pm$3.92 & \\
\midrule
\multirow{2}{*}{Context Trace 2} & \textbf{-355} & -388 & -377 (A2C) & \multirow{2}{*}{-376 (DDQN)} \\
 & \textbf{$\pm$2.57} & $\pm$7.33 & $\pm$8.05 & \\
\midrule
\multirow{2}{*}{Context Trace 3} & \textbf{-240} & -271 & -262 (CLEAR) & \multirow{2}{*}{-203 (SAC)} \\
 & \textbf{$\pm$4.41} & $\pm$21.9 & $\pm$2.2 &  \\
\midrule
\multirow{2}{*}{Context Trace 4} & \textbf{-378} & -395 & -399 (A2C) & \multirow{2}{*}{-357 (SAC)} \\
 & \textbf{$\pm$3.02} & $\pm$9.32 & $\pm$7.76 & \\

\bottomrule
\end{tabular}
\end{table*}

We compare \sys and \sys-P in \Cref{table:windypend_lcpo_vars}. We tuned $\kappa$ on the Pendulum-v1 environment, similar to all baselines, and finalized on $\kappa=10$. Despite this, \sys-P fail to outperform the best baseline for each context trace even on Pendulum-v1. The stark contrast between \sys and \sys-P is due to the difficulty in tuning $\kappa$. \sys's optimization constraint guarantees that the policy is anchored on past contexts, while the loss term added in \sys-P motivates for this change to be small. Although the setup in \sys-P is the Lagrangian dual of the setup in \sys, they are only equivalent when $\kappa$ is tuned properly per each gradient step.
\section{Locomotion tasks in Gymnasium}
\label{sec:pend_env}

\begin{figure}[ht]
    \centering
    \includegraphics[width=\linewidth]{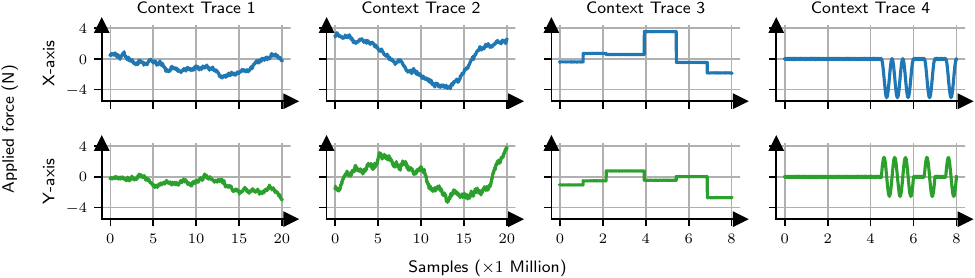}
    \caption{External wind force, per axis and context trace.}
    \label{fig:windypend_traces}
\end{figure}

\subsection{Full results}

\Cref{table:gym_all_p1,table:gym_all_p2} presents the lifelong return for all agents, environments and context traces.
\Cref{table:gym_sigma_all} presents the lifelong return for multiple \sys agents with different \gls{ood} thresholds $\sigma$.
\Cref{table:gym_buffer_all} presents the lifelong return for multiple \sys agents with different buffer sizes $n_b$.

\begin{table}[ht]
\caption{Average and 95th percentile confidence ranges for lifelong return for different algorithms and conditions in environments with external wind processes. Schemes with superiority beyond 95\% confidence are highlighted in bold. \textbf{Continued in \Cref{table:gym_all_p2}.}}
\label{table:gym_all_p1}
\scriptsize
\centering

\begin{tabular}{c l c c c c c c c c c c c}
\toprule
& & \multicolumn{6}{c}{\textbf{Online Learning}} & & \multicolumn{4}{c}{\textbf{Prescient Policies}} \\
\cmidrule{3-8} \cmidrule{10-13}
& & LCPO & Online EWC & Sliding OGD & CLEAR & MBCD & MBPO & & A2C & TRPO & DDQN & SAC \\

\midrule

\multirow{10}{*}{\rotatebox[origin=c]{90}{Pendulum}} & \multirow{2}{*}{Trace 1} & -190 & -204 & -1195 & -215 & -207 & -325 &  & \multirow{2}{*}{-208} & \multirow{2}{*}{-197} & \multirow{2}{*}{-188} & \multirow{2}{*}{\textbf{-187}} \\
 &  & $\pm$0.45 & $\pm$2.37 & $\pm$4.38 & $\pm$0.42 & $\pm$23.4 & $\pm$47.2 &  &  &  &  &  \\
\cmidrule{2-13}
 & \multirow{2}{*}{Trace 2} & \textbf{-355} & -525 & -1232 & -429 & -577 & -843 &  & \multirow{2}{*}{-407} & \multirow{2}{*}{-399} & \multirow{2}{*}{\textbf{-376}} & \multirow{2}{*}{-380} \\
 &  & \textbf{$\pm$2.57} & $\pm$22.6 & $\pm$3.25 & $\pm$1.58 & $\pm$71.3 & $\pm$24.2 &  &  &  &  &  \\
\cmidrule{2-13}
 & \multirow{2}{*}{Trace 3} & \textbf{-240} & -345 & -1181 & -262 & -349 & -603 &  & \multirow{2}{*}{-224} & \multirow{2}{*}{-320} & \multirow{2}{*}{-228} & \multirow{2}{*}{\textbf{-203}} \\
 &  & \textbf{$\pm$4.41} & $\pm$29.6 & $\pm$3.78 & $\pm$2.20 & $\pm$28.9 & $\pm$213 &  &  &  &  &  \\
\cmidrule{2-13}
 & \multirow{2}{*}{Trace 4} & \textbf{-378} & -412 & -1233 & -411 & -430 & -553 &  & \multirow{2}{*}{-383} & \multirow{2}{*}{-381} & \multirow{2}{*}{-384} & \multirow{2}{*}{\textbf{-357}} \\
 &  & \textbf{$\pm$3.02} & $\pm$6.52 & $\pm$4.41 & $\pm$2.45 & $\pm$15.4 & $\pm$90.9 &  &  &  &  &  \\

\midrule

\multirow{10}{*}{\rotatebox[origin=c]{90}{Inverse Pendulum}} & \multirow{2}{*}{Trace 1} & 18.7 & 3.47 & 2.21 & 2.28 & 6.17 & 2.88 &  & \multirow{2}{*}{24.4} & \multirow{2}{*}{\textbf{25.7}} & \multirow{2}{*}{24.1} & \multirow{2}{*}{24.8} \\
 &  & $\pm$2.85 & $\pm$0.32 & $\pm$0.01 & $\pm$0.10 & $\pm$6.10 & $\pm$1.78 &  &  &  &  &  \\
\cmidrule{2-13}
 & \multirow{2}{*}{Trace 2} & \textbf{5.31} & 1.55 & 1.58 & 1.60 & 2.10 & 1.52 &  & \multirow{2}{*}{7.86} & \multirow{2}{*}{\textbf{8.51}} & \multirow{2}{*}{7.18} & \multirow{2}{*}{7.75} \\
 &  & \textbf{$\pm$0.54} & $\pm$0.03 & $\pm$0.00 & $\pm$0.03 & $\pm$1.06 & $\pm$0.01 &  &  &  &  &  \\
\cmidrule{2-13}
 & \multirow{2}{*}{Trace 3} & 15.4 & 5.56 & 4.24 & 2.45 & 14.2 & 14.4 &  & \multirow{2}{*}{15.5} & \multirow{2}{*}{\textbf{16.5}} & \multirow{2}{*}{15.1} & \multirow{2}{*}{14.3} \\
 &  & $\pm$0.22 & $\pm$0.88 & $\pm$0.02 & $\pm$0.36 & $\pm$0.77 & $\pm$0.23 &  &  &  &  &  \\
\cmidrule{2-13}
 & \multirow{2}{*}{Trace 4} & 92.4 & 5.43 & 4.64 & 2.87 & 46.5 & 37.6 &  & \multirow{2}{*}{102} & \multirow{2}{*}{\textbf{112}} & \multirow{2}{*}{53.7} & \multirow{2}{*}{40.1} \\
 &  & $\pm$0.19 & $\pm$0.88 & $\pm$0.01 & $\pm$0.17 & $\pm$4.85 & $\pm$0.98 &  &  &  &  &  \\

\midrule

\multirow{10}{*}{\rotatebox[origin=c]{90}{Inverse Double Pendulum}} & \multirow{2}{*}{Trace 1} & 114 & 38.8 & 38.5 & 34.2 & 77.3 & 68.5 &  & \multirow{2}{*}{142} & \multirow{2}{*}{\textbf{165}} & \multirow{2}{*}{88.0} & \multirow{2}{*}{95.5} \\
 &  & $\pm$3.54 & $\pm$2.15 & $\pm$0.10 & $\pm$0.46 & $\pm$13.5 & $\pm$19.3 &  &  &  &  &  \\
\cmidrule{2-13}
 & \multirow{2}{*}{Trace 2} & 56.2 & 30.1 & 30.9 & 28.9 & 44.2 & 37.2 &  & \multirow{2}{*}{51.4} & \multirow{2}{*}{\textbf{64.0}} & \multirow{2}{*}{46.8} & \multirow{2}{*}{49.5} \\
 &  & $\pm$2.28 & $\pm$1.03 & $\pm$0.03 & $\pm$0.20 & $\pm$3.27 & $\pm$9.92 &  &  &  &  &  \\
\cmidrule{2-13}
 & \multirow{2}{*}{Trace 3} & 65.1 & 36.1 & 36.4 & 27.1 & 45.8 & 42.7 &  & \multirow{2}{*}{63.5} & \multirow{2}{*}{\textbf{74.0}} & \multirow{2}{*}{63.6} & \multirow{2}{*}{47.4} \\
 &  & $\pm$0.98 & $\pm$1.80 & $\pm$0.06 & $\pm$0.66 & $\pm$2.38 & $\pm$1.55 &  &  &  &  &  \\
\cmidrule{2-13}
 & \multirow{2}{*}{Trace 4} & 89.2 & 37.7 & 41.4 & 29.3 & 75.0 & 59.5 &  & \multirow{2}{*}{91.9} & \multirow{2}{*}{\textbf{96.9}} & \multirow{2}{*}{90.7} & \multirow{2}{*}{78.0} \\
 &  & $\pm$0.33 & $\pm$3.22 & $\pm$0.07 & $\pm$0.80 & $\pm$4.65 & $\pm$2.86 &  &  &  &  &  \\

\midrule

\multirow{10}{*}{\rotatebox[origin=c]{90}{Hopper}} & \multirow{2}{*}{Trace 1} & \textbf{244} & 34.1 & 11.4 & 3.92 & 153 & 130 &  & \multirow{2}{*}{\textbf{240}} & \multirow{2}{*}{215} & \multirow{2}{*}{155} & \multirow{2}{*}{184} \\
 &  & \textbf{$\pm$5.58} & $\pm$5.05 & $\pm$0.42 & $\pm$0.54 & $\pm$33.2 & $\pm$12.8 &  &  &  &  &  \\
\cmidrule{2-13}
 & \multirow{2}{*}{Trace 2} & \textbf{144} & 28.2 & 12.0 & 2.87 & 76.0 & 66.0 &  & \multirow{2}{*}{128} & \multirow{2}{*}{\textbf{131}} & \multirow{2}{*}{86.6} & \multirow{2}{*}{96.7} \\
 &  & \textbf{$\pm$3.44} & $\pm$4.98 & $\pm$0.53 & $\pm$0.25 & $\pm$15.3 & $\pm$18.2 &  &  &  &  &  \\
\cmidrule{2-13}
 & \multirow{2}{*}{Trace 3} & \textbf{508} & 27.0 & 19.3 & 3.62 & 240 & 264 &  & \multirow{2}{*}{\textbf{520}} & \multirow{2}{*}{492} & \multirow{2}{*}{327} & \multirow{2}{*}{308} \\
 &  & \textbf{$\pm$2.54} & $\pm$10.4 & $\pm$0.60 & $\pm$0.31 & $\pm$81.7 & $\pm$49.9 &  &  &  &  &  \\
\cmidrule{2-13}
 & \multirow{2}{*}{Trace 4} & \textbf{296} & 22.2 & 12.8 & 2.94 & 127 & 205 &  & \multirow{2}{*}{\textbf{319}} & \multirow{2}{*}{297} & \multirow{2}{*}{189} & \multirow{2}{*}{189} \\
 &  & \textbf{$\pm$2.59} & $\pm$8.47 & $\pm$0.38 & $\pm$0.10 & $\pm$27.3 & $\pm$37.5 &  &  &  &  &  \\

\midrule

\multirow{10}{*}{\rotatebox[origin=c]{90}{Reacher}} & \multirow{2}{*}{Trace 1} & -26.5 & -16.3 & -6622 & -1780 & -169 & -1862 &  & \multirow{2}{*}{-8.00} & \multirow{2}{*}{-7.59} & \multirow{2}{*}{\textbf{-7.59}} & \multirow{2}{*}{-14.3} \\
 &  & $\pm$0.21 & $\pm$0.60 & $\pm$124 & $\pm$958 & $\pm$122 & $\pm$2255 &  &  &  &  &  \\
\cmidrule{2-13}
 & \multirow{2}{*}{Trace 2} & -19.8 & -16.4 & -6733 & -1695 & -339 & -4308 &  & \multirow{2}{*}{-8.33} & \multirow{2}{*}{-3866} & \multirow{2}{*}{\textbf{-7.41}} & \multirow{2}{*}{-15.3} \\
 &  & $\pm$0.11 & $\pm$0.37 & $\pm$74.0 & $\pm$731 & $\pm$493 & $\pm$3764 &  &  &  &  &  \\
\cmidrule{2-13}
 & \multirow{2}{*}{Trace 3} & -27.3 & -15.4 & -6722 & -799 & -596 & -2058 &  & \multirow{2}{*}{-9.63} & \multirow{2}{*}{-8.42} & \multirow{2}{*}{\textbf{-8.41}} & \multirow{2}{*}{-15.7} \\
 &  & $\pm$0.89 & $\pm$0.23 & $\pm$131 & $\pm$161 & $\pm$857 & $\pm$4173 &  &  &  &  &  \\
\cmidrule{2-13}
 & \multirow{2}{*}{Trace 4} & -29.4 & -16.6 & -6718 & -1392 & -121 & -2354 &  & \multirow{2}{*}{-9.63} & \multirow{2}{*}{-8.57} & \multirow{2}{*}{\textbf{-7.97}} & \multirow{2}{*}{-15.7} \\
 &  & $\pm$0.29 & $\pm$0.44 & $\pm$102 & $\pm$1060 & $\pm$161 & $\pm$2988 &  &  &  &  &  \\
\bottomrule
\end{tabular}

\end{table}

\begin{table}[ht]
\caption{Average and 95th percentile confidence ranges for lifelong return for different algorithms and conditions in environments with external wind processes. Schemes with superiority beyond 95\% confidence are highlighted in bold. \textbf{Continued from \Cref{table:gym_all_p1}.}}
\label{table:gym_all_p2}
\scriptsize
\centering

\begin{tabular}{c l c c c c c c c c c c c c}
\toprule
& & \multicolumn{7}{c}{\textbf{Online Learning}} & & \multicolumn{4}{c}{\textbf{Prescient Policies}} \\
\cmidrule{3-9} \cmidrule{11-14}
& & LCPO & A2C & TRPO & BFDQN & PT-DQN & DDQN & SAC & & A2C & TRPO & DDQN & SAC \\

\midrule

\multirow{10}{*}{\rotatebox[origin=c]{90}{Pendulum}} & \multirow{2}{*}{Trace 1} & -190 & -201 & -222 & -346 & -205 & -219 & -207 &  & \multirow{2}{*}{-208} & \multirow{2}{*}{-197} & \multirow{2}{*}{-188} & \multirow{2}{*}{\textbf{-187}} \\
 &  & $\pm$0.45 & $\pm$3.92 & $\pm$4.14 & $\pm$15.0 & $\pm$2.80 & $\pm$8.88 & $\pm$4.12 &  &  &  &  &  \\
\cmidrule{2-14}
 & \multirow{2}{*}{Trace 2} & \textbf{-355} & -377 & -548 & -709 & -835 & -756 & -636 &  & \multirow{2}{*}{-407} & \multirow{2}{*}{-399} & \multirow{2}{*}{\textbf{-376}} & \multirow{2}{*}{-380} \\
 &  & \textbf{$\pm$2.57} & $\pm$8.05 & $\pm$41.3 & $\pm$64.9 & $\pm$120 & $\pm$103 & $\pm$62.3 &  &  &  &  &  \\
\cmidrule{2-14}
 & \multirow{2}{*}{Trace 3} & \textbf{-240} & -307 & -511 & -766 & -596 & -612 & -402 &  & \multirow{2}{*}{-224} & \multirow{2}{*}{-320} & \multirow{2}{*}{-228} & \multirow{2}{*}{\textbf{-203}} \\
 &  & \textbf{$\pm$4.41} & $\pm$32.6 & $\pm$63.4 & $\pm$75.1 & $\pm$79.5 & $\pm$76.2 & $\pm$31.0 &  &  &  &  &  \\
\cmidrule{2-14}
 & \multirow{2}{*}{Trace 4} & \textbf{-378} & -399 & -648 & -693 & -436 & -409 & -418 &  & \multirow{2}{*}{-383} & \multirow{2}{*}{-381} & \multirow{2}{*}{-384} & \multirow{2}{*}{\textbf{-357}} \\
 &  & \textbf{$\pm$3.02} & $\pm$7.76 & $\pm$31.4 & $\pm$53.3 & $\pm$32.0 & $\pm$7.59 & $\pm$5.52 &  &  &  &  &  \\

\midrule

\multirow{10}{*}{\rotatebox[origin=c]{90}{Inverse Pendulum}} & \multirow{2}{*}{Trace 1} & 18.7 & 7.57 & 4.61 & 5.79 & 13.6 & 12.7 & 3.97 &  & \multirow{2}{*}{24.4} & \multirow{2}{*}{\textbf{25.7}} & \multirow{2}{*}{24.1} & \multirow{2}{*}{24.8} \\
 &  & $\pm$2.85 & $\pm$2.31 & $\pm$1.84 & $\pm$1.13 & $\pm$2.27 & $\pm$1.38 & $\pm$0.67 &  &  &  &  &  \\
\cmidrule{2-14}
 & \multirow{2}{*}{Trace 2} & \textbf{5.31} & 2.36 & 1.73 & 1.74 & 2.24 & 1.60 & 1.59 &  & \multirow{2}{*}{7.86} & \multirow{2}{*}{\textbf{8.51}} & \multirow{2}{*}{7.18} & \multirow{2}{*}{7.75} \\
 &  & \textbf{$\pm$0.54} & $\pm$0.45 & $\pm$0.28 & $\pm$0.09 & $\pm$0.13 & $\pm$0.06 & $\pm$0.06 &  &  &  &  &  \\
\cmidrule{2-14}
 & \multirow{2}{*}{Trace 3} & 15.4 & 15.0 & 7.94 & 4.28 & 15.0 & 14.9 & 14.7 &  & \multirow{2}{*}{15.5} & \multirow{2}{*}{\textbf{16.5}} & \multirow{2}{*}{15.1} & \multirow{2}{*}{14.3} \\
 &  & $\pm$0.22 & $\pm$0.46 & $\pm$2.36 & $\pm$1.58 & $\pm$0.17 & $\pm$0.14 & $\pm$0.05 &  &  &  &  &  \\
\cmidrule{2-14}
 & \multirow{2}{*}{Trace 4} & 92.4 & \textbf{93.7} & 75.9 & 79.6 & 75.5 & 92.1 & 50.5 &  & \multirow{2}{*}{102} & \multirow{2}{*}{\textbf{112}} & \multirow{2}{*}{53.7} & \multirow{2}{*}{40.1} \\
 &  & $\pm$0.19 & \textbf{$\pm$0.21} & $\pm$10.6 & $\pm$2.88 & $\pm$3.18 & $\pm$1.18 & $\pm$1.46 &  &  &  &  &  \\

\midrule

\multirow{10}{*}{\rotatebox[origin=c]{90}{Inverse Double Pendulum}} & \multirow{2}{*}{Trace 1} & 114 & 118 & 86.3 & 48.7 & 76.6 & 80.0 & 75.6 &  & \multirow{2}{*}{142} & \multirow{2}{*}{\textbf{165}} & \multirow{2}{*}{88.0} & \multirow{2}{*}{95.5} \\
 &  & $\pm$3.54 & $\pm$9.47 & $\pm$13.9 & $\pm$7.99 & $\pm$1.91 & $\pm$0.95 & $\pm$5.67 &  &  &  &  &  \\
\cmidrule{2-14}
 & \multirow{2}{*}{Trace 2} & 56.2 & 54.3 & 30.0 & 25.6 & 43.6 & 42.4 & 38.8 &  & \multirow{2}{*}{51.4} & \multirow{2}{*}{\textbf{64.0}} & \multirow{2}{*}{46.8} & \multirow{2}{*}{49.5} \\
 &  & $\pm$2.28 & $\pm$1.72 & $\pm$3.60 & $\pm$0.08 & $\pm$1.52 & $\pm$1.17 & $\pm$2.37 &  &  &  &  &  \\
\cmidrule{2-14}
 & \multirow{2}{*}{Trace 3} & 65.1 & 63.4 & 49.7 & 39.0 & 61.9 & 61.6 & 46.8 &  & \multirow{2}{*}{63.5} & \multirow{2}{*}{\textbf{74.0}} & \multirow{2}{*}{63.6} & \multirow{2}{*}{47.4} \\
 &  & $\pm$0.98 & $\pm$2.37 & $\pm$3.93 & $\pm$4.49 & $\pm$0.29 & $\pm$0.26 & $\pm$0.23 &  &  &  &  &  \\
\cmidrule{2-14}
 & \multirow{2}{*}{Trace 4} & 89.2 & 87.9 & 75.9 & 67.6 & 93.9 & 93.9 & 78.7 &  & \multirow{2}{*}{91.9} & \multirow{2}{*}{\textbf{96.9}} & \multirow{2}{*}{90.7} & \multirow{2}{*}{78.0} \\
 &  & $\pm$0.33 & $\pm$0.53 & $\pm$1.85 & $\pm$4.37 & $\pm$0.16 & $\pm$0.18 & $\pm$0.63 &  &  &  &  &  \\

\midrule

\multirow{10}{*}{\rotatebox[origin=c]{90}{Hopper}} & \multirow{2}{*}{Trace 1} & \textbf{244} & 203 & 164 & 93.5 & 86.0 & 136 & 154 &  & \multirow{2}{*}{\textbf{240}} & \multirow{2}{*}{215} & \multirow{2}{*}{155} & \multirow{2}{*}{184} \\
 &  & \textbf{$\pm$5.58} & $\pm$7.73 & $\pm$7.25 & $\pm$3.24 & $\pm$13.9 & $\pm$8.83 & $\pm$14.1 &  &  &  &  &  \\
\cmidrule{2-14}
 & \multirow{2}{*}{Trace 2} & \textbf{144} & 97.9 & 65.8 & 23.2 & 30.0 & 68.1 & 81.7 &  & \multirow{2}{*}{128} & \multirow{2}{*}{\textbf{131}} & \multirow{2}{*}{86.6} & \multirow{2}{*}{96.7} \\
 &  & \textbf{$\pm$3.44} & $\pm$3.07 & $\pm$9.11 & $\pm$6.85 & $\pm$7.43 & $\pm$4.11 & $\pm$5.69 &  &  &  &  &  \\
\cmidrule{2-14}
 & \multirow{2}{*}{Trace 3} & \textbf{508} & 484 & 443 & 137 & 52.2 & 220 & 360 &  & \multirow{2}{*}{\textbf{520}} & \multirow{2}{*}{492} & \multirow{2}{*}{327} & \multirow{2}{*}{308} \\
 &  & \textbf{$\pm$2.54} & $\pm$13.8 & $\pm$17.1 & $\pm$12.2 & $\pm$6.63 & $\pm$6.15 & $\pm$39.2 &  &  &  &  &  \\
\cmidrule{2-14}
 & \multirow{2}{*}{Trace 4} & \textbf{296} & 263 & 186 & 105 & 117 & 161 & 258 &  & \multirow{2}{*}{\textbf{319}} & \multirow{2}{*}{297} & \multirow{2}{*}{189} & \multirow{2}{*}{189} \\
 &  & \textbf{$\pm$2.59} & $\pm$5.59 & $\pm$14.9 & $\pm$6.14 & $\pm$13.7 & $\pm$8.02 & $\pm$14.4 &  &  &  &  &  \\

\midrule

\multirow{10}{*}{\rotatebox[origin=c]{90}{Reacher}} & \multirow{2}{*}{Trace 1} & -26.5 & -118 & -2346 & -9.41 & -2492 & -7.84 & -370 &  & \multirow{2}{*}{-8.00} & \multirow{2}{*}{-7.59} & \multirow{2}{*}{\textbf{-7.59}} & \multirow{2}{*}{-14.3} \\
 &  & $\pm$0.21 & $\pm$45.9 & $\pm$471 & $\pm$0.15 & $\pm$1898 & $\pm$0.03 & $\pm$169 &  &  &  &  &  \\
\cmidrule{2-14}
 & \multirow{2}{*}{Trace 2} & -19.8 & -102 & -4728 & -11.5 & -19.9 & -7.78 & -618 &  & \multirow{2}{*}{-8.33} & \multirow{2}{*}{-3866} & \multirow{2}{*}{\textbf{-7.41}} & \multirow{2}{*}{-15.3} \\
 &  & $\pm$0.11 & $\pm$34.6 & $\pm$515 & $\pm$3.52 & $\pm$17.5 & $\pm$0.07 & $\pm$290 &  &  &  &  &  \\
\cmidrule{2-14}
 & \multirow{2}{*}{Trace 3} & -27.3 & -133 & -71.7 & -10.1 & -5916 & -8.56 & -258 &  & \multirow{2}{*}{-9.63} & \multirow{2}{*}{-8.42} & \multirow{2}{*}{\textbf{-8.41}} & \multirow{2}{*}{-15.7} \\
 &  & $\pm$0.89 & $\pm$61.5 & $\pm$80.0 & $\pm$0.32 & $\pm$2335 & $\pm$0.04 & $\pm$140 &  &  &  &  &  \\
\cmidrule{2-14}
 & \multirow{2}{*}{Trace 4} & -29.4 & -117 & -1780 & -9.17 & -23.2 & -9.21 & -157 &  & \multirow{2}{*}{-9.63} & \multirow{2}{*}{-8.57} & \multirow{2}{*}{\textbf{-7.97}} & \multirow{2}{*}{-15.7} \\
 &  & $\pm$0.29 & $\pm$56.3 & $\pm$272 & $\pm$0.45 & $\pm$10.4 & $\pm$1.22 & $\pm$59.6 &  &  &  &  &  \\
\bottomrule
\end{tabular}

\end{table}
\begin{table*}[t!]
\caption{Average and 95th percentile confidence ranges for lifelong return in \sys with different \gls{ood} metrics and thresholds and other agents for different conditions in environments with external wind processes. L2 stands for the L2-distance \gls{ood} metric and MHD stands for the Mahalanobis distance \gls{ood} metric. Schemes with superiority beyond 95\% confidence are highlighted in bold (\sys agents are only compared against baselines).}
\label{table:gym_sigma_all}
\scriptsize
\centering

\begin{tabular}{c l@{} c c c c c c @{}c c c c@{}}
\toprule
& & \multicolumn{5}{c}{\textbf{LCPO (L2)}} & & \multicolumn{2}{c}{\textbf{LCPO (L2)}} & & \\
\cmidrule{3-7}
\cmidrule{9-10}
& & $\sigma^2=0.25$ & $\sigma^2=0.5$ & $\sigma^2=1$ & $\sigma^2=2$ & $\sigma^2=4$ &  & $\sigma_{\text{MHD}}^2=6$ & $\sigma_{\text{MHD}}^2=12$ & Best Baseline & Best Prescient \\

\midrule

\multirow{10}{*}{\rotatebox[origin=c]{90}{Pendulum}} & \multirow{2}{*}{Trace 1} & -189 & -190 & -190 & -194 & -199 & & -188 & \textbf{-188} & -201 (A2C) & \multirow{2}{*}{-187 (SAC)} \\
 &  & $\pm$0.52 & $\pm$0.66 & $\pm$0.45 & $\pm$0.62 & $\pm$0.32 &  & $\pm$0.50 & \textbf{$\pm$0.53} & $\pm$3.92 &  \\
\cmidrule{2-12}
 & \multirow{2}{*}{Trace 2} & -349 & -354 & -355 & -359 & -361 & & \textbf{-346} & -346 & -377 (A2C) & \multirow{2}{*}{-376 (DDQN)} \\
 &  & $\pm$2.12 & $\pm$2.68 & $\pm$2.57 & $\pm$2.64 & $\pm$3.03 &  & \textbf{$\pm$1.66} & $\pm$2.49 & $\pm$8.05 &  \\
\cmidrule{2-12}
 & \multirow{2}{*}{Trace 3} & -242 & -242 & -240 & -235 & \textbf{-231} & & -233 & -233 & -262 (CLEAR) & \multirow{2}{*}{-203 (SAC)} \\
 &  & $\pm$5.34 & $\pm$5.34 & $\pm$4.41 & $\pm$3.94 & \textbf{$\pm$4.13} &  & $\pm$4.39 & $\pm$4.40 & $\pm$2.20 &  \\
\cmidrule{2-12}
 & \multirow{2}{*}{Trace 4} & -375 & -377 & -378 & -380 & -385 & & \textbf{-369} & -373 & -399 (A2C) & \multirow{2}{*}{-357 (SAC)} \\
 &  & $\pm$2.76 & $\pm$4.12 & $\pm$3.02 & $\pm$2.99 & $\pm$2.78 &  & \textbf{$\pm$3.45} & $\pm$3.07 & $\pm$7.76 &  \\

\midrule

\multirow{10}{*}{\rotatebox[origin=c]{90}{Inverse Pendulum}} & \multirow{2}{*}{Trace 1} & \textbf{21.5} & 18.9 & 18.7 & 6.05 & 2.77 & & 4.54 & 17.1 & 13.6 (PT-DQN) & \multirow{2}{*}{25.7 (TRPO)} \\
 &  & \textbf{$\pm$1.72} & $\pm$3.03 & $\pm$2.85 & $\pm$2.13 & $\pm$0.16 &  & $\pm$2.64 & $\pm$2.96 & $\pm$2.27 &  \\
\cmidrule{2-12}
 & \multirow{2}{*}{Trace 2} & 4.85 & 4.24 & 5.31 & \textbf{5.35} & 2.79 & & 5.12 & 5.28 & 2.36 (A2C) & \multirow{2}{*}{8.51 (TRPO)} \\
 &  & $\pm$0.60 & $\pm$0.71 & $\pm$0.54 & \textbf{$\pm$0.56} & $\pm$0.71 &  & $\pm$0.68 & $\pm$0.62 & $\pm$0.45 &  \\
\cmidrule{2-12}
 & \multirow{2}{*}{Trace 3} & 15.1 & 15.6 & 15.4 & 15.4 & 15.4 & & \textbf{16.3} & 16.3 & 15.0 (A2C) & \multirow{2}{*}{16.5 (TRPO)} \\
 &  & $\pm$0.36 & $\pm$0.13 & $\pm$0.22 & $\pm$0.23 & $\pm$0.23 &  & \textbf{$\pm$0.03} & $\pm$0.04 & $\pm$0.46 &  \\
\cmidrule{2-12}
 & \multirow{2}{*}{Trace 4} & 94.5 & 92.4 & 92.4 & 92.4 & 92.4 & & \textbf{100} & 97.6 & 93.7 (A2C) & \multirow{2}{*}{112 (TRPO)} \\
 &  & $\pm$0.24 & $\pm$0.19 & $\pm$0.19 & $\pm$0.19 & $\pm$0.19 &  & \textbf{$\pm$0.25} & $\pm$0.39 & $\pm$0.21 &  \\

\midrule

\multirow{10}{*}{\rotatebox[origin=c]{90}{Inverse Double Pendulum}} & \multirow{2}{*}{Trace 1} & 118 & 118 & 114 & 118 & 118 & & \textbf{223} & 221 & 118 (A2C) & \multirow{2}{*}{165 (TRPO)} \\
 &  & $\pm$5.26 & $\pm$5.26 & $\pm$3.54 & $\pm$5.26 & $\pm$5.26 &  & \textbf{$\pm$5.39} & $\pm$6.53 & $\pm$9.47 &  \\
\cmidrule{2-12}
 & \multirow{2}{*}{Trace 2} & 53.4 & 54.2 & 56.2 & 54.9 & 54.9 & & 62.9 & \textbf{63.3} & 54.3 (A2C) & \multirow{2}{*}{64.0 (TRPO)} \\
 &  & $\pm$3.35 & $\pm$1.60 & $\pm$2.28 & $\pm$2.23 & $\pm$2.23 &  & $\pm$2.95 & \textbf{$\pm$4.44} & $\pm$1.72 &  \\
\cmidrule{2-12}
 & \multirow{2}{*}{Trace 3} & 65.5 & 65.0 & 65.1 & 64.9 & 64.9 & & 62.5 & 64.5 & 63.4 (A2C) & \multirow{2}{*}{74.0 (TRPO)} \\
 &  & $\pm$0.92 & $\pm$0.91 & $\pm$0.98 & $\pm$0.91 & $\pm$0.91 &  & $\pm$4.07 & $\pm$3.41 & $\pm$2.37 &  \\
\cmidrule{2-12}
 & \multirow{2}{*}{Trace 4} & 89.9 & 89.4 & 89.2 & 89.4 & 89.4 & & 90.1 & 89.7 & \textbf{93.9} (DDQN) & \multirow{2}{*}{96.9 (TRPO)} \\
 &  & $\pm$0.27 & $\pm$0.24 & $\pm$0.33 & $\pm$0.24 & $\pm$0.24 &  & $\pm$0.13 & $\pm$0.14 & \textbf{$\pm$0.18} &  \\

\midrule

\multirow{10}{*}{\rotatebox[origin=c]{90}{Hopper}} & \multirow{2}{*}{Trace 1} & 247 & 250 & 244 & 235 & 222 & & 245 & \textbf{252} & 203 (A2C) & \multirow{2}{*}{240 (A2C)} \\
 &  & $\pm$6.96 & $\pm$8.38 & $\pm$5.58 & $\pm$2.81 & $\pm$3.49 &  & $\pm$10.2 & \textbf{$\pm$8.11} & $\pm$7.73 &  \\
\cmidrule{2-12}
 & \multirow{2}{*}{Trace 2} & 144 & 144 & 144 & 139 & 138 & & \textbf{147} & 146 & 97.9 (A2C) & \multirow{2}{*}{131 (TRPO)} \\
 &  & $\pm$3.48 & $\pm$3.20 & $\pm$3.44 & $\pm$3.14 & $\pm$3.50 &  & \textbf{$\pm$3.59} & $\pm$3.79 & $\pm$3.07 &  \\
\cmidrule{2-12}
 & \multirow{2}{*}{Trace 3} & 520 & 512 & 508 & 508 & 508 & & \textbf{539} & 533 & 484 (A2C) & \multirow{2}{*}{520 (A2C)} \\
 &  & $\pm$4.01 & $\pm$2.46 & $\pm$2.54 & $\pm$2.46 & $\pm$2.46 &  & \textbf{$\pm$8.78} & $\pm$6.40 & $\pm$13.8 &  \\
\cmidrule{2-12}
 & \multirow{2}{*}{Trace 4} & \textbf{302} & 298 & 296 & 292 & 291 & & 300 & 298 & 263 (A2C) & \multirow{2}{*}{319 (A2C)} \\
 &  & \textbf{$\pm$3.26} & $\pm$3.22 & $\pm$2.59 & $\pm$2.04 & $\pm$4.09 &  & $\pm$3.98 & $\pm$2.88 & $\pm$5.59 &  \\

\midrule

\multirow{10}{*}{\rotatebox[origin=c]{90}{Reacher}} & \multirow{2}{*}{Trace 1} & -26.7 & -26.5 & -26.5 & -24.5 & -21.6 & & -25.1 & -25.1 & \textbf{-7.84} (DDQN) & \multirow{2}{*}{-7.59 (DDQN)} \\
 &  & $\pm$0.18 & $\pm$0.19 & $\pm$0.21 & $\pm$0.32 & $\pm$0.19 &  & $\pm$0.12 & $\pm$0.09 & \textbf{$\pm$0.03} &  \\
\cmidrule{2-12}
 & \multirow{2}{*}{Trace 2} & -25.5 & -22.6 & -19.8 & -19.8 & -19.8 & & -23.9 & -23.8 & \textbf{-7.78} (DDQN) & \multirow{2}{*}{-7.41 (DDQN)} \\
 &  & $\pm$0.11 & $\pm$0.13 & $\pm$0.11 & $\pm$0.12 & $\pm$0.12 &  & $\pm$0.23 & $\pm$0.22 & \textbf{$\pm$0.07} &  \\
\cmidrule{2-12}
 & \multirow{2}{*}{Trace 3} & -27.0 & -27.0 & -27.3 & -29.0 & -28.4 & & -25.4 & -25.5 & \textbf{-8.56} (DDQN) & \multirow{2}{*}{-8.41 (DDQN)} \\
 &  & $\pm$1.07 & $\pm$1.07 & $\pm$0.89 & $\pm$0.62 & $\pm$0.67 &  & $\pm$0.82 & $\pm$0.56 & \textbf{$\pm$0.04} &  \\
\cmidrule{2-12}
 & \multirow{2}{*}{Trace 4} & -30.6 & -30.4 & -29.4 & -27.7 & -26.0 & & -24.6 & -24.2 & \textbf{-9.17} (BFDQN) & \multirow{2}{*}{-7.97 (DDQN)} \\
 &  & $\pm$0.29 & $\pm$0.36 & $\pm$0.29 & $\pm$0.37 & $\pm$0.34 &  & $\pm$0.19 & $\pm$0.23 & \textbf{$\pm$0.45} &  \\
\bottomrule
\end{tabular}

\end{table*}
\begin{table*}[ht]
\caption{Average and 95th percentile confidence ranges for lifelong return in \sys with multiple buffer sizes and other agents for different conditions in environments with external wind processes. Schemes with superiority beyond 95\% confidence are highlighted in bold (\sys agents are only compared against baselines).}
\label{table:gym_buffer_all}
\scriptsize
\centering
\begin{tabular}{c l c c c c c c}
\toprule
& & \multicolumn{4}{c}{\textbf{LCPO}} & & \\
\cmidrule{3-6}
& & $n_b=20M$ & $n_b=200K$ & $n_b=500$ & $n_b=25$ & Best Baseline & Best Prescient \\

\midrule

\multirow{10}{*}{\rotatebox[origin=c]{90}{Pendulum}} & \multirow{2}{*}{Trace 1} & -191 & \textbf{-190} & -190 & -193 & -201 (A2C) & \multirow{2}{*}{-187 (SAC)} \\
 &  & $\pm$0.65 & \textbf{$\pm$0.45} & $\pm$0.62 & $\pm$0.99 & $\pm$3.92 &  \\
\cmidrule{2-8}
 & \multirow{2}{*}{Trace 2} & -355 & \textbf{-355} & -356 & -366 & -377 (A2C) & \multirow{2}{*}{-376 (DDQN)} \\
 &  & $\pm$2.02 & \textbf{$\pm$2.57} & $\pm$2.01 & $\pm$3.57 & $\pm$8.05 &  \\
\cmidrule{2-8}
 & \multirow{2}{*}{Trace 3} & \textbf{-240} & -240 & -240 & -259 & -262 (CLEAR) & \multirow{2}{*}{-203 (SAC)} \\
 &  & \textbf{$\pm$6.83} & $\pm$4.41 & $\pm$9.10 & $\pm$10.7 & $\pm$2.20 &  \\
\cmidrule{2-8}
 & \multirow{2}{*}{Trace 4} & -379 & \textbf{-378} & -379 & -400 & -399 (A2C) & \multirow{2}{*}{-357 (SAC)} \\
 &  & $\pm$3.28 & \textbf{$\pm$3.02} & $\pm$3.78 & $\pm$4.62 & $\pm$7.76 &  \\

\midrule

\multirow{10}{*}{\rotatebox[origin=c]{90}{Inverse Pendulum}} & \multirow{2}{*}{Trace 1} & 18.4 & 18.7 & \textbf{20.8} & 14.7 & 13.6 (PT-DQN) & \multirow{2}{*}{25.7 (TRPO)} \\
 &  & $\pm$2.54 & $\pm$2.85 & \textbf{$\pm$1.72} & $\pm$3.48 & $\pm$2.27 &  \\
\cmidrule{2-8}
 & \multirow{2}{*}{Trace 2} & 4.69 & \textbf{5.31} & 4.91 & 4.61 & 2.36 (A2C) & \multirow{2}{*}{8.51 (TRPO)} \\
 &  & $\pm$0.66 & \textbf{$\pm$0.54} & $\pm$0.63 & $\pm$0.77 & $\pm$0.45 &  \\
\cmidrule{2-8}
 & \multirow{2}{*}{Trace 3} & 15.4 & 15.4 & 15.4 & 15.4 & 15.0 (A2C) & \multirow{2}{*}{16.5 (TRPO)} \\
 &  & $\pm$0.23 & $\pm$0.22 & $\pm$0.22 & $\pm$0.23 & $\pm$0.46 &  \\
\cmidrule{2-8}
 & \multirow{2}{*}{Trace 4} & 92.3 & 92.4 & 92.4 & 92.4 & \textbf{93.7} (A2C) & \multirow{2}{*}{112 (TRPO)} \\
 &  & $\pm$0.22 & $\pm$0.19 & $\pm$0.19 & $\pm$0.19 & \textbf{$\pm$0.21} &  \\

\midrule

\multirow{10}{*}{\rotatebox[origin=c]{90}{Inverse Double Pendulum}} & \multirow{2}{*}{Trace 1} & 117 & 114 & 117 & 118 & 118 (A2C) & \multirow{2}{*}{165 (TRPO)} \\
 &  & $\pm$5.30 & $\pm$3.54 & $\pm$5.29 & $\pm$5.26 & $\pm$9.47 &  \\
\cmidrule{2-8}
 & \multirow{2}{*}{Trace 2} & 54.6 & 56.2 & 54.7 & 54.9 & 54.3 (A2C) & \multirow{2}{*}{64.0 (TRPO)} \\
 &  & $\pm$2.19 & $\pm$2.28 & $\pm$2.19 & $\pm$2.23 & $\pm$1.72 &  \\
\cmidrule{2-8}
 & \multirow{2}{*}{Trace 3} & 64.9 & 65.1 & 65.0 & 64.9 & 63.4 (A2C) & \multirow{2}{*}{74.0 (TRPO)} \\
 &  & $\pm$0.90 & $\pm$0.98 & $\pm$0.92 & $\pm$0.91 & $\pm$2.37 &  \\
\cmidrule{2-8}
 & \multirow{2}{*}{Trace 4} & 89.4 & 89.2 & 89.4 & 89.4 & \textbf{93.9} (DDQN) & \multirow{2}{*}{96.9 (TRPO)} \\
 &  & $\pm$0.24 & $\pm$0.33 & $\pm$0.24 & $\pm$0.24 & \textbf{$\pm$0.18} &  \\

\midrule

\multirow{10}{*}{\rotatebox[origin=c]{90}{Hopper}} & \multirow{2}{*}{Trace 1} & 238 & \textbf{244} & 241 & 233 & 203 (A2C) & \multirow{2}{*}{240 (A2C)} \\
 &  & $\pm$4.45 & \textbf{$\pm$5.58} & $\pm$5.37 & $\pm$3.22 & $\pm$7.73 &  \\
\cmidrule{2-8}
 & \multirow{2}{*}{Trace 2} & 141 & \textbf{144} & 139 & 135 & 97.9 (A2C) & \multirow{2}{*}{131 (TRPO)} \\
 &  & $\pm$4.13 & \textbf{$\pm$3.44} & $\pm$3.76 & $\pm$2.54 & $\pm$3.07 &  \\
\cmidrule{2-8}
 & \multirow{2}{*}{Trace 3} & \textbf{509} & 508 & 509 & 508 & 484 (A2C) & \multirow{2}{*}{520 (A2C)} \\
 &  & \textbf{$\pm$2.48} & $\pm$2.54 & $\pm$2.42 & $\pm$2.46 & $\pm$13.8 &  \\
\cmidrule{2-8}
 & \multirow{2}{*}{Trace 4} & \textbf{296} & 296 & 291 & 279 & 263 (A2C) & \multirow{2}{*}{319 (A2C)} \\
 &  & \textbf{$\pm$2.48} & $\pm$2.59 & $\pm$4.23 & $\pm$5.53 & $\pm$5.59 &  \\

\midrule

\multirow{10}{*}{\rotatebox[origin=c]{90}{Reacher}} & \multirow{2}{*}{Trace 1} & -26.6 & -26.5 & -26.5 & -26.4 & \textbf{-7.84} (DDQN) & \multirow{2}{*}{-7.59 (DDQN)} \\
 &  & $\pm$0.15 & $\pm$0.21 & $\pm$0.19 & $\pm$0.26 & \textbf{$\pm$0.03} &  \\
\cmidrule{2-8}
 & \multirow{2}{*}{Trace 2} & -19.8 & -19.8 & -19.8 & -19.8 & \textbf{-7.78} (DDQN) & \multirow{2}{*}{-7.41 (DDQN)} \\
 &  & $\pm$0.12 & $\pm$0.11 & $\pm$0.12 & $\pm$0.12 & \textbf{$\pm$0.07} &  \\
\cmidrule{2-8}
 & \multirow{2}{*}{Trace 3} & -27.8 & -27.3 & -27.8 & -28.1 & \textbf{-8.56} (DDQN) & \multirow{2}{*}{-8.41 (DDQN)} \\
 &  & $\pm$0.84 & $\pm$0.89 & $\pm$0.97 & $\pm$0.53 & \textbf{$\pm$0.04} &  \\
\cmidrule{2-8}
 & \multirow{2}{*}{Trace 4} & -29.5 & -29.4 & -29.4 & -29.3 & \textbf{-9.17} (BFDQN) & \multirow{2}{*}{-7.97 (DDQN)} \\
 &  & $\pm$0.31 & $\pm$0.29 & $\pm$0.73 & $\pm$0.83 & \textbf{$\pm$0.45} &  \\

\bottomrule
\end{tabular}
\end{table*}

\subsection{Computational Load}
\label{sec:gym_runtime}

\begin{table*}[ht]
\caption{Average and 95\% confidence interval for training time per environment interactions, in $\mu$sec.}
\label{table:windy_time}
\scriptsize
\centering

\begin{tabular}{c c c c c c c c c c c c c}
\toprule
\multirow{2}{*}{LCPO} & \multirow{2}{*}{A2C} & \multirow{2}{*}{TRPO} & \multirow{2}{*}{DDQN} & \multirow{2}{*}{SAC} & Sliding & \multirow{2}{*}{CLEAR} & \multirow{2}{*}{BFDQN} & \multirow{2}{*}{MBPO} & \multirow{2}{*}{MBCD} & Online & \multirow{2}{*}{PT-DQN} \\
& & & & & OGD & & & & & EWC & \\
\midrule
0.78 & 0.52 & 0.45 & 0.45 & 0.50 & 0.65 & 0.43 & 0.31 & 2.19 & 5.62 & 3.97 & 0.42 \\
$\pm$0.03 & $\pm$0.03 & $\pm$0.03 & $\pm$0.03 & $\pm$0.02 & $\pm$0.01 & $\pm$0.01 & $\pm$0.01 & $\pm$0.04 & $\pm$0.32 & $\pm$0.02 & $\pm$0.01 \\
\bottomrule
\end{tabular}

\end{table*}

\sys is about 1.5$\times$ more computationally demanding than the leading baseline \gls{a2c}. \Cref{table:windy_time} depicts the total runtime per each environment interaction in the experiments in \S\ref{sec:all_eval}. 

\subsection{Action Space}
\label{sec:pend_discretize}

Pendulum-v1 and Mujoco environments by default have continuous action spaces. We observed instability while learning policies with continuous policy classes even for the prescient policies, and were concerned about how this can affect the validity of our online experiments, which are considerably more challenging. As the action space is tangent to our problem, we discretized each dimension of the action space to $15$ atoms, spaced equally from the minimum to the maximum action in each dimension. This stabilized training greatly, and is not surprising, as past work~\citep{tang2019discretizing} supports this observation. The reward metric, continuous state space and truncation and termination conditions remain unchanged.

We provide the achieved episodic return for all baselines in \S\ref{sec:eval} in \Cref{table:pend_original}, over 10 seeds for the Pendulum-v1 envioronment, which we can compare to SB3~\citep{stable-baselines3} and RL-Zoo~\citep{rl-zoo3} reported figures. These experiments finished in approximately 46 minutes. \gls{a2c}, \gls{ddqn} and \gls{sac} were trained for 8000 epochs, and \gls{trpo} was trained for 500 epochs. Evaluations are on 1000 episodes. As these results show, the agents exhibit stable training with a discretized action space.

\begin{table}[ht]
\caption{Average episodic returns and 95th percentile confidence ranges for different algorithms in the Pendulum-v1 environment with discretized and continuous action space.}
\label{table:pend_original}
\scriptsize
\centering
\begin{tabular}{l c c c c c}
\toprule
Episodic Return & A2C & TRPO & DDQN & SAC & MBPO \\
\midrule
Discrete & -165+-5  & -166+-8  & -149+-2  & -146+-2  & -161+-3 \\
\midrule
SB3~\citep{stable-baselines3} + RL-Zoo~\citep{rl-zoo3} & -203  & -224  & ---  & -176 & --- \\
\bottomrule
\end{tabular}
\end{table}

\subsection{Experiment Setup}
\label{sec:pend_train_params}

We use Gymnasium (v0.29.1, MIT license) and Mujoco (v3.1.1, Apache-2.0 license). Our baseline and \sys implementations use the Pytorch~\citep{pytorch} (v1.13.1, BSD-style license) library. 
\Cref{table:windy_train} is a comprehensive list of all hyperparameters used in training and the environment.

All baselines were tuned on Pendulum-v1 via a multi-phased grid search, similar to that in \S\ref{subsec:base_ewc}. General parameters such as discount horizon were copied from the base RL algorithm each baseline is using (e.g., online EWC is using SAC, and copied SAC-specific parameters directly). Several \sys hyperparameters were copied from \gls{trpo}, \gls{sac} and \gls{a2c} (namely, entropy target, entropy learning rate, damping coefficient, rollout length, $\lambda$, $\gamma$) and the rest ($c_{anchor}$, $c_{recent}$ and base entropy) were tuned with an informal search with a separate context trace (not in the evaluation set) in Pendulum-v1. The \gls{ood} threshold $\sigma$ was not tuned with a search.

{\scriptsize
\begin{longtable}{c l l}
\caption{Training setup and hyperparameters for gymnasium environments with external wind.} \label{table:windy_train} \\
\toprule
\textbf{Group} & \textbf{Hyperparameter} & \textbf{Value} \\
\midrule
\endfirsthead

\toprule
\textbf{Group} & \textbf{Hyperparameter} & \textbf{Value} \\
\midrule
\endhead

\endfoot

\endlastfoot

\multirow{8}{*}{Neural network} & Hidden layers & (64, 64) \\
\cmidrule{2-3}
& Hidden layer activation & Relu \\
\cmidrule{2-3}
& Output layer activation & Actors: Softmax, Critics and \gls{ddqn}: Identity mapping \\
\cmidrule{2-3}
& Optimizer & Adam ($\beta_1=0.9$, $\beta_2=0.999$) \citep{kingma2017adam} \\
\cmidrule{2-3}
& Learning rate & Actor: 0.0004, Critic and \gls{ddqn}: 0.001 \\
\cmidrule{2-3}
& Weight decay & $10^{-4}$ \\
\midrule

\multirow{4}{*}{\gls{rl} training (general)} & \multirow{2}{*}{Random seeds} & 25 in main experiments (\S\ref{sec:all_eval}), except for MBPO and MBCD \\
& & 5 seeds in ablations (\S\ref{sec:ood_agg_eval}, \S\ref{sec:ood_size_eval}, \S\ref{sec:lcpo_p}, \S\ref{subsec:base_mbpo}, \S\ref{subsec:base_ewc}) \\
\cmidrule{2-3}
& $\lambda$ (for \gls{gae} in \gls{a2c} and \gls{trpo}) & 0.9 \\
\cmidrule{2-3}
& $\gamma$ & 0.99 \\
\midrule

\gls{a2c} & Rollout per epoch & 200 \\
\midrule

\multirow{4}{*}{\gls{trpo}} & Rollout per epoch & 3200 \\
\cmidrule{2-3}
& Damping coefficient & 0.1 \\
\cmidrule{2-3}
& Stepsize & 0.01 \\
\midrule

\multirow{9}{*}{\gls{ddqn}} & Rollout per epoch & 200 \\
\cmidrule{2-3}
& Batch Size & 512 \\
\cmidrule{2-3}
& Initial fully random period & 1000 epochs \\
\cmidrule{2-3}
& $\epsilon$-greedy schedule & 1 to 0 in 5000 epochs \\
\cmidrule{2-3}
& Polyak $\alpha$ & 0.01 \\
\cmidrule{2-3}
& Buffer size $N$ & All samples ($N=20M$ or $N=8M$) \\
\midrule

\multirow{11}{*}{\gls{sac}} & Rollout per epoch & 200 \\
\cmidrule{2-3}
& Batch Size & 512 \\
\cmidrule{2-3}
& Initial fully random period & 1000 epochs \\
\cmidrule{2-3}
& Base Entropy & 0.1 \\
\cmidrule{2-3}
& Entropy Target & $0.1 \ln(15)$ \\
\cmidrule{2-3}
& Log-Entropy Learning Rate & 1e-3 \\
\cmidrule{2-3}
& Polyak $\alpha$ & 0.01 \\
\cmidrule{2-3}
& Buffer size $N$ & All samples ($N=20M$ or $N=8M$) \\
\midrule

\multirow{12}{*}{\sys} & Rollout per epoch & 200 \\
\cmidrule{2-3}
& Base Entropy & 0.03 \\
\cmidrule{2-3}
& Entropy Target & $0.1 \ln(15)$ \\
\cmidrule{2-3}
& Log-Entropy Learning Rate & 1e-3 \\
\cmidrule{2-3}
& Buffer Size $n_b$ & 1\% of samples ($200K$ or $80K$) \\
\cmidrule{2-3}
& Damping coefficient & 0.1 \\
\cmidrule{2-3}
& $c_{anchor}$ & 0.0001 \\
\cmidrule{2-3}
& $c_{recent}$ & 0.1 \\
\cmidrule{2-3}
& $\sigma$ & 1 \\
\midrule

\multirow{5}{*}{\sys-P} & PPO Clipping $\epsilon$ & 0.2 \\
\cmidrule{2-3}
& PPO Iterations (Max) & 30 \\
\cmidrule{2-3}
& PPO Max KL & 0.01 \\
\cmidrule{2-3}
& $\kappa$ & 10 \\
\midrule

\multirow{5}{*}{\gls{mbcd}} & h & 1000 (default was 100/300) \\
\cmidrule{2-3}
& max\_std & 3 (default was 0.5) \\
\cmidrule{2-3}
& N (ensemble size) & 5 \\
\cmidrule{2-3}
& NN hidden layers & (64, 64, 64) \\
\midrule

\multirow{5}{*}{\gls{mbpo}} & M (model rollouts) & 512 \\
\cmidrule{2-3}
& N (ensemble size) & 5 \\
\cmidrule{2-3}
& k (rollout length) & 1 \\
\cmidrule{2-3}
& G (gradient steps) & 1 \\
\midrule

\multirow{3}{*}{Online \gls{ewc}} & averaging weight $\beta$ & 0.00007 (equivalent to $\sim3M$ samples at rollout=200) \\
\cmidrule{2-3}
& scaling factor $\alpha$ & 0.1 \\
\midrule

\multirow{3}{*}{Sliding OGD} & learning rate $\alpha$ & 1e-4 \\
\cmidrule{2-3}
& N (window size) & 1000 episodes \\
\midrule

\multirow{7}{*}{CLEAR} & Value Clone Coefficient & 1e-2 \\
\cmidrule{2-3}
& Policy Clone Coefficient & 1e-3 \\
\cmidrule{2-3}
& Entropy Coefficient & 5e-3 \\
\cmidrule{2-3}
& V-Trace $\rho$ & 1 \\
\cmidrule{2-3}
& V-Trace c & 1 \\
\midrule

\multirow{3}{*}{BFDQN} & Depth & 13 \\
\cmidrule{2-3}
& Benna Fusi Buffer Length & 2000 \\
\cmidrule{2-3}
& $g_{1\_2}$ & 1e-3 \\
\midrule


\multirow{7}{*}{PT-DQN} & Permanent Learning Rate & 1e-5 \\
\cmidrule{2-3}
& Transient Learning Rate & 1e-3 \\
\cmidrule{2-3}
& Target Update Period $N$ & 2000 steps \\
\cmidrule{2-3}
& Permanent Update period $K$ & 20000 steps \\
\cmidrule{2-3}
& Transient Forget Factor $\lambda$ & 0.999 \\

\bottomrule
\end{longtable}
}
\section{Straggler mitigation}
\label{sec:lbalance_env}

\begin{figure}[ht]
    \centering
    \begin{subfigure}{0.47\linewidth}
        \includegraphics[trim=11cm 0.25cm 1.5cm 2.25cm, page=1, clip, width=\linewidth]{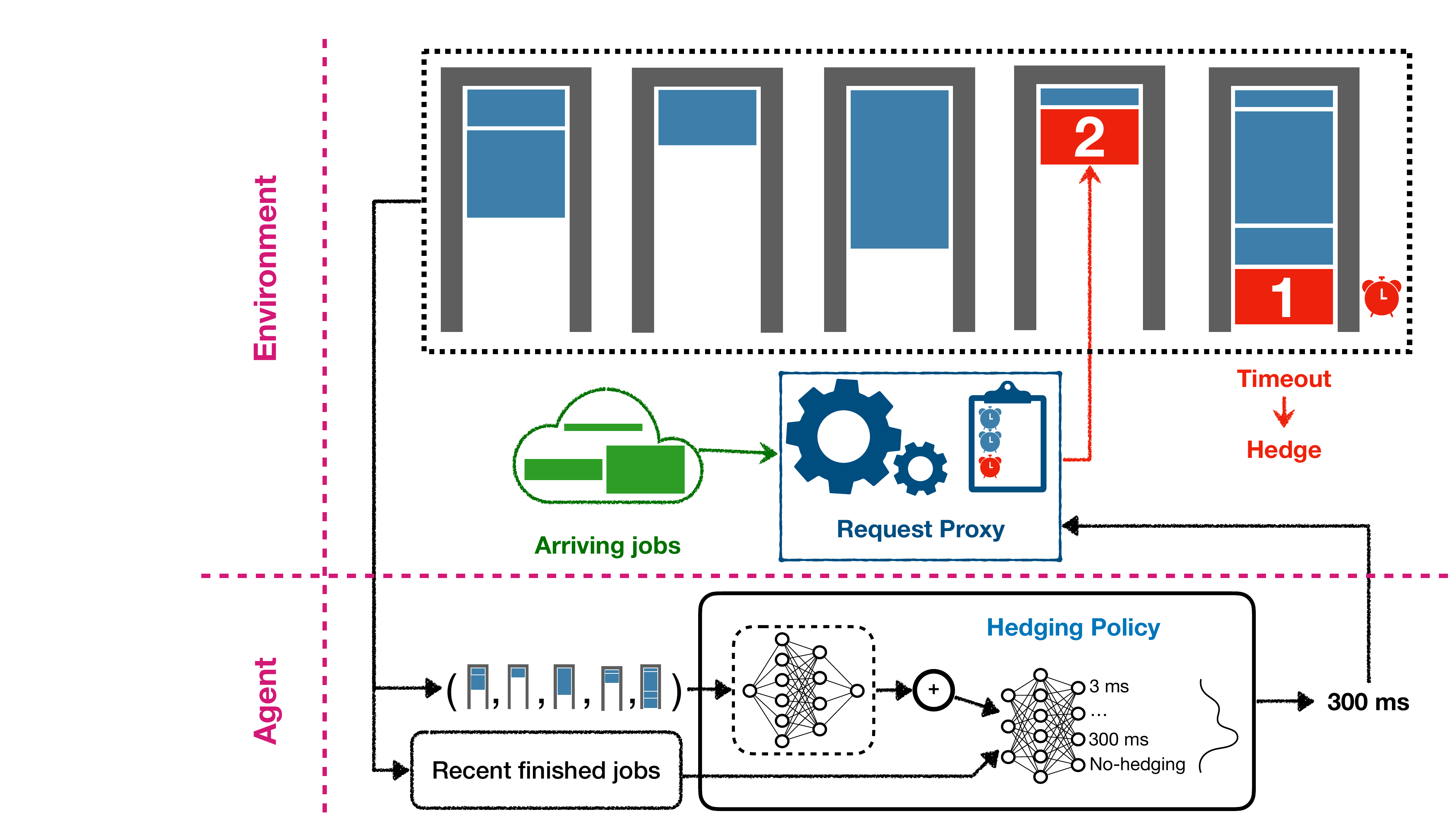}
        \caption{Illustration of a request proxy with hedging.}
        \label{fig:lbalance_fig}
    \end{subfigure}
    \hspace{0.1in}
    \begin{subfigure}{0.47\linewidth}
        \includegraphics[width=\linewidth]{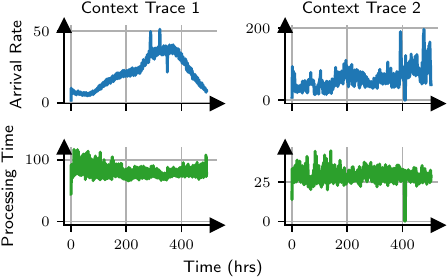}
        \caption{Request arrival rate and processing time per input.}
        \label{fig:lbalance_traces}
    \end{subfigure}
    \caption{Request arrival rate and processing time per input.}
    \label{fig:lbalance_figures}
\end{figure}

\subsection{Full results}

\begin{figure}[ht]
    \centering
    \includegraphics[width=\linewidth]{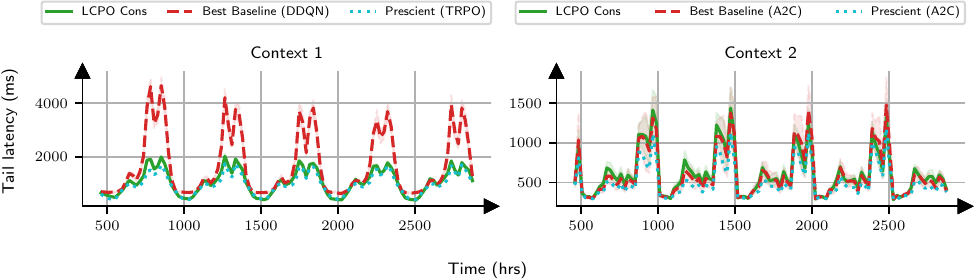}
    \caption{Tail latency with 95th percentile confidence intervals as training progresses (lower is better). We consider an initial learning period of 3.5 million samples. \sys remains close to the prescient throughout contexts, while baselines suffer from non-stationarity.}
    \label{fig:lbalance_results_full}
\end{figure}

\Cref{fig:lbalance_results_full} plots the tail latency across experiment time in the straggler mitigation environment for both contexts.

\subsection{Experiment Setup}
\label{sec:lbalance_train_params}

We use the straggler mitigation environment from prior work~\citep{hamadanian2022demistify}, with a similar configuration except with 9 actions (timeouts of $600^{ms}$ and $1000^{ms}$ added). Similar to \S\ref{sec:pend_train_params}, our implementations of baselines and \sys use the Pytorch~\citep{pytorch} (v1.13.1, BSD-style license) library. The environment code and dataset is not public and was released to us with a proprietary license.
\Cref{table:lbalance_train} is a comprehensive list of all hyperparameters used in training and the environment.

All baselines were tuned on a separate workload. \sys hyperparameters were copied from the gymnasium experiments, except for base entropy which was tuned with an informal search with a separate workload (not in the evaluation set).

{\scriptsize
\begin{longtable}{c l l}
\caption{Training setup and hyperparameters for straggler mitigation experiments.}\label{table:lbalance_train}\\
\toprule
\textbf{Group} & \textbf{Hyperparameter} & \textbf{Value} \\
\midrule
\endfirsthead

\toprule
\textbf{Group} & \textbf{Hyperparameter} & \textbf{Value} \\
\midrule
\endhead

\endfoot

\endlastfoot

\multirow{10}{*}{Neural network} & \multirow{2}{*}{Hidden layers} & $\phi$ network: (32, 16) \\
\cmidrule{3-3}
& & $\rho$ network: (32, 32) \\
\cmidrule{2-3}
& Hidden layer activation function & Relu \\
\cmidrule{2-3}
& Output layer activation function & Actors: Softmax, Critics and \gls{ddqn}: Identity mapping \\
\cmidrule{2-3}
& Optimizer & Adam ($\beta_1=0.9$, $\beta_2=0.999$) \citep{kingma2017adam} \\
\cmidrule{2-3}
& Learning rate & 0.001 \\
\cmidrule{2-3}
& Weight decay & $10^{-4}$ \\
\midrule

\multirow{4}{*}{\gls{rl} training (general)} & Random seeds & 10 \\
\cmidrule{2-3}
& $\lambda$ (for \gls{gae} in \gls{a2c} and \gls{trpo}) & 0.95 \\
\cmidrule{2-3}
& $\gamma$ & 0.9 \\
\midrule

\gls{a2c} & Rollout per epoch & 4608 \\
\midrule

\multirow{4}{*}{\gls{trpo}} & Rollout per epoch & 10240 \\
\cmidrule{2-3}
& Damping coefficient & 0.1 \\
\cmidrule{2-3}
& Stepsize & 0.01 \\
\midrule

\multirow{9}{*}{\gls{ddqn}} & Rollout per epoch & 128 \\
\cmidrule{2-3}
& Batch Size & 512 \\
\cmidrule{2-3}
& Initial fully random period & 1000 epochs \\
\cmidrule{2-3}
& $\epsilon$-greedy schedule & 1 to 0 in 5000 epochs \\
\cmidrule{2-3}
& Polyak $\alpha$ & 0.01 \\
\cmidrule{2-3}
& Buffer size $N$ & All samples ($N=21M$) \\
\midrule

\\
\\
\\
\\
\\
\\

\multirow{11}{*}{\gls{sac}} & Rollout per epoch & 128 \\
\cmidrule{2-3}
& Batch Size & 512 \\
\cmidrule{2-3}
& Initial fully random period & 1000 epochs \\
\cmidrule{2-3}
& Base Entropy & 0.01 \\
\cmidrule{2-3}
& Entropy Target & $0.1 \ln(9)$ \\
\cmidrule{2-3}
& Log-Entropy Learning Rate & 1e-3 \\
\cmidrule{2-3}
& Polyak $\alpha$ & 0.005 \\
\cmidrule{2-3}
& Buffer size $N$ & All samples ($N=21M$) \\
\midrule

\multirow{12}{*}{\sys} & Rollout per epoch & 128 \\
\cmidrule{2-3}
& Base Entropy & 0.01 \\
\cmidrule{2-3}
& Entropy Target & $0.1 \ln(9)$ \\
\cmidrule{2-3}
& Log-Entropy Learning Rate & 1e-3 \\
\cmidrule{2-3}
& Buffer Size $n_b$ & $210K$ \\
\cmidrule{2-3}
& Damping coefficient & 0.1 \\
\cmidrule{2-3}
& $c_{anchor}$ & 0.0001 \\
\cmidrule{2-3}
& $c_{recent}$ & 0.1 \\
\midrule

\multirow{5}{*}{\gls{mbcd}} & h & 300000 (default was 100/300) \\
\cmidrule{2-3}
& max\_std & 3 (default was 0.5) \\
\cmidrule{2-3}
& N (ensemble size) & 5 \\
\cmidrule{2-3}
& NN hidden layers & (64, 64, 64) \\
\midrule

\multirow{5}{*}{\gls{mbpo}} & M (model rollouts) & 512 \\
\cmidrule{2-3}
& N (ensemble size) & 5 \\
\cmidrule{2-3}
& k (rollout length) & 1 \\
\cmidrule{2-3}
& G (gradient steps) & 1 \\
\midrule

\multirow{3}{*}{Online \gls{ewc}} & averaging weight $\beta$ & 0.00007 (equivalent to $\sim2M$ samples at rollout=128) \\
\cmidrule{2-3}
& scaling factor $\alpha$ & 0.1 \\

\bottomrule

\end{longtable}
}
\end{appendices}

\end{document}